\documentclass[lettersize,journal]{IEEEtran}
\usepackage{amsmath,amsfonts}
\usepackage{algorithmic}
\usepackage{algorithm}
\usepackage{array}
\usepackage[caption=false,font=normalsize,labelfont=sf,textfont=sf]{subfig}
\usepackage{textcomp}
\usepackage{stfloats}
\usepackage{url}
\usepackage{verbatim}
\usepackage{graphicx}
\usepackage{cite}
\usepackage[table]{xcolor}
\usepackage{ulem}
\usepackage{booktabs}       % professional-quality tables
\usepackage{amsfonts}       % blackboard math symbols
\usepackage{nicefrac}       % compact symbols for 1/2, etc.
\usepackage{microtype}      % microtypography
\usepackage[dvipsnames]{xcolor}         % colors
\usepackage{enumitem}
\usepackage{multirow}
\usepackage{amsmath}
\usepackage{mdframed}
\usepackage{subcaption}
\usepackage{soul}
\begin{document}

\title{MotionGPT-2: A General-Purpose Motion-Language Model for Motion Generation and Understanding}

\author{%
\textbf{Yuan Wang}$^{1,5\ddagger}$ 
~~
\textbf{Di Huang}$^{2\ddagger}$ 
~~
\textbf{Yaqi Zhang}$^{3}$  
~~
\textbf{Wanli Ouyang}$^{4, 5}$,~\IEEEmembership{Senior Member,~IEEE} \\
~~
\textbf{Jile Jiao}$^{1,6}$ 
~~
\textbf{Xuetao Feng}$^{1,7}$ 
~~ 
\textbf{Dan Xu}$^{8}$,~\IEEEmembership{Member,~IEEE}
~~
\textbf{Shixiang Tang}$^{4\clubsuit}$  
~~
\\
% ~~
% \textbf{Yan Zhou}$^{8}$ 
% ~~
% \textbf{Pengfei Wan}$^{8}$  \\
$^{1}$Tsinghua University
~~
$^{2}$The University of Sydney
~~
$^{3}$University of Science and Technology of China \\
~~
$^{4}$The Chinese University of Hong Kong 
~~
$^{5}$Shanghai Artificial Intelligence Laboratory \\
~~
$^{6}$Intime Department Store 
~~
$^{7}$Deepeleph
~~
$^{8}$HKUST \quad
$^{\ddagger}$Equal Contribution 
~~
$^{\clubsuit}$Corresponding Author
~~
        % <-this % stops a space
\thanks{This paper was produced by the IEEE Publication Technology Group. They are in Piscataway, NJ.}% <-this % stops a space
\thanks{Manuscript received April 19, 2021; revised August 16, 2021.}}

\markboth{Journal of \LaTeX\ Class Files,~Vol.~14, No.~8, August~2021}%
{Shell \MakeLowercase{\textit{et al.}}: A Sample Article Using IEEEtran.cls for IEEE Journals}


% \maketitle

\twocolumn[{
\renewcommand\twocolumn[1][]{#1}%
\maketitle
\begin{center}
  \centering
  \vspace{-1em}
  \captionsetup{type=figure}
  \includegraphics[width=\linewidth]{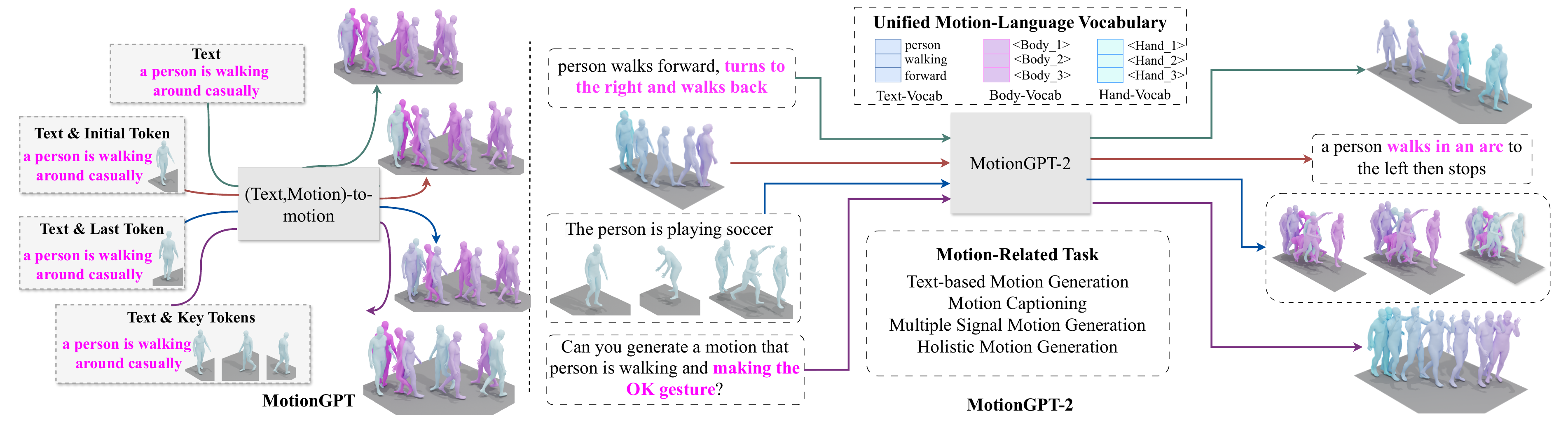}
  \captionof{figure}{
     {This paper proposes a versatile motion-language framework via fine-tuned LLMs given different instructions, named MotionGPT-2. Compared with the previous MotionGPT~\cite{zhang2023motiongpt}, our MotionGPT-2 not only retains the unique capability of accommodating multiple control conditions, but also solve various motion-related tasks using a unified model. }}
  \label{fig:teaser}
  \vspace{1.5em}
\end{center}%
\vspace{-0.4cm}
}]

\begin{abstract}
Generating lifelike human motions from descriptive texts has experienced remarkable research focus in recent
years, propelled by the emerging requirements of digital humans.
Despite impressive advances, existing approaches are often constrained by limited control modalities, task specificity, and focus solely on body motion representations.
% , restricting their versatility and applicability in creative industries. 
% In this paper, we propose a \textbf{Motion} \textbf{G}eneral-\textbf{P}urpose genera\textbf{T}or (MotionGPT-2), a unified Large Motion-Language Model (LMLM) that can handle multiple motion-relevant tasks and accept multi-modal control conditions utilizing pre-trained Large Language Model (LLMs).
% In this paper, we introduce MotionGPT-2, a unified Large Motion-Language Model (LMLM). MotionGPT-2 is designed to address these limitations by accommodating multiple motion-relevant tasks and supporting multimodal control conditions through pre-trained Large Language Models (LLMs).
In this paper, we present MotionGPT-2, a unified Large Motion-Language Model (LMLM) that addresses these limitations. MotionGPT-2 accommodates multiple motion-relevant tasks and supports multimodal control conditions through pre-trained Large Language Models (LLMs).
% Specifically, we first quantize the multi-modal control signals, \textit{e.g.,} text and single-frame poses, into discrete LLM-interpretable tokens and extend the motion-aware tokens into the original LLM's vocabulary. 
% We then formulate multi-modal tokens in a unified prompt instruction, directing the LLMs to generate the motion answer in a pretraining-then-finetuning paradigm. Most notably, our proposed MotionGPT-2 demonstrates a unified motion-language model with multimodal control signals by tuning a mere 1\% of LLM parameters. 
% Our approach begins by quantizing multi-modal control inputs—such as text and single-frame poses—into discrete, LLM-interpretable tokens and incorporating them into the LLM's vocabulary. These tokens are then structured into unified prompt instructions, enabling the LLM to generate motion outputs within a pretraining-then-finetuning paradigm.
It quantizes multimodal inputs—such as text and single-frame poses—into discrete, LLM-interpretable tokens, seamlessly integrating them into the LLM's vocabulary. These tokens are then organized into unified prompts, guiding the LLM to generate motion outputs through a pretraining-then-finetuning paradigm. We also show that the proposed MotionGPT-2 is highly adaptable to the challenging 3D holistic motion generation task, enabled by the innovative motion discretization framework, Part-Aware VQVAE, which ensures fine-grained representations of body and hand movements. 
% We also show that the proposed MotionGPT-2 is highly adaptable to the challenging 3D holistic motion generation task, enabled by the innovative motion discretization framework, Part-Aware VQVAE, which ensures fine-grained representations of body and hand movements. 
% To address the complexities of 3D holistic motion generation, we introduce a novel motion discretization framework, Part-Aware VQVAE, which captures fine-grained representations of both body and hand movements.
% Extensive experiments and visualizations validate the effectiveness of our method. In the text-based motion generation task, our generalized MotionGPT-2 yields 0.496 Top1 accuracy and 0.191 FID on HumanML3D, as well as 0.427 Top1 accuracy and 0.614 FID on KIT-ML, comparable to the specific diffusion-based models.
% Comprehensive experiments and visualizations demonstrate the effectiveness of our method.
Extensive experiments and visualizations validate the effectiveness of our method, demonstrating the adaptability of MotionGPT-2 across motion generation, motion captioning, and generalized motion completion tasks. 
\end{abstract}


\vspace{-0.6cm}
\section{Introduction}

\IEEEPARstart{H}{uman} motion~\cite{ding2022towards,yu2023toward,chen2023spatiotemporal,xia20223d,li2021survey,zhou2023dual,liu2023multi} plays a pivotal role in various applications, including video gaming, robotic assistance, and VR/AR. 
Recent years have witnessed a rapid
development of Generative Artificial Intelligence (GenAI)~\cite{yuan2019bridge,yang2025spatio,ren2023hr,ma2021spatial,liang2023layout,wang2025uniadapter,hu2023dear}, paving the way for novel methods to motion synthesis.
% Various control conditions, such as textual descriptions~\cite{zhang2023t2m,chen2023mld,zhang2022motiondiffuse}, music pieces~\cite{li2021ai,li2022danceformer,lee2019dancing}, and human poses~\cite{zhu2023motionbert}, provide an intuitive and user-friendly approach to animating virtual characters or controlling humanoid robots. 
% Among these multi-modal control conditions, descriptive texts have emerged as a dominant research frontier, due to their ease of derivation from language descriptors, developing a simplified user interface to interact with computers~\cite{ahuja2019language2pose,wang2024holistic,petrovich2022temos,zhang2022motiondiffuse}. 
% prior researches have expanded beyond motion generation to other motion-related tasks like motion captioning~\cite{tevet2022motionclip,guo2022tm2t} and motion prediction~\cite{ma2022multi,yuan2020dlow,zhang2021we}. 
% \IEEEpubidadjcol
% 词组

Despite the impressive performance achieved by existing motion generation methods~\cite{petrovich2022temos,zhang2022motiondiffuse,tevet2022human,zhuang2022music2dance}, 
% their limitations lie in their focus on a single type of control condition, the task-specific frameworks, and body-only motion representations. 
current methods suffer from three major limitations:
\textit{(1) Limited control conditions}. 
Some text-to-motion approaches typically target only a single type of control condition, \textit{i.e.}, either textual descriptions or multiple frame poses. This narrow scope constrains their practical applications in scenarios that require the generation of motion sequences conditioned on both text descriptions and multiple key-frame human poses, \textit{e.g.}, social robotics and film animation. Although methods in \cite{cohan2024flexible,karunratanakul2023guided,xie2023omnicontrol,diller2024cg} aim to integrate additional control modalities, they treat control signals as separate modalities, neglecting to construct a unified multi-modal representation.
% Most existing methods \hd{TODO: cite} rely on a single control input, such as text or key-frame poses, reducing their applicability in scenarios requiring mixed-modal control, e.g., combining text descriptions with multiple human poses.
\textit{(2) Task-specific frameworks without general world knowledge}. 
% Present studies have not fully exploited the general world knowledge of Large Language Models (LLMs) to address multiple motion-related tasks within a unified framework.
% Previous methods~\cite{zhang2023t2m,chen2023executing,zhang2024motiondiffuse,guo2024momask} rely on task-specific frameworks (\textit{e.g.}, diffusion-based and GPT-based), constraining adaptability to diverse motion-related tasks and hardly generalize effectively to unseen data. 
% Recently, several works~\cite{zhang2023motiongpt,jiang2023motiongpt,luo2024m,wu2024motionllm,feng2024chatpose,lin2024chathuman} have initiated the unification of motion-related task within a single framework. Employing a encoder-decoder language model, MotionGPT~\cite{jiang2023motiongpt} and M3-GPT~\cite{luo2024m} propose a versatile motion-language model that synthesizes realistic human motions and descriptive text via prompt-driven instructions. However, these approaches have not yet explored the full scope of LLMs in understanding and generating human motion. Recent studies~\cite{feng2024chatpose,lin2024chathuman} confirms that LLMs are exceptionally capable of interpreting and understanding 3D human motions through comprehensive world knowledge. Therefore, it is important to unleash the general-purpose reasoning ability of LLMs for motion-related tasks, as demonstrated in other vision tasks. 
Existing models are often task-specific, such as diffusion-based and GPT-based frameworks~\cite{zhang2023t2m, chen2023executing,zhang2022motiondiffuse,guo2024momask}. These methods lack the adaptability needed for multiple tasks (e.g., captioning, in-betweening, prediction) and cannot fully leverage the world knowledge embedded in Large Language Models (LLMs). While recent efforts~\cite{jiang2023motiongpt, luo2024m} present unified frameworks, their reliance on relatively small language models means the broader potential of LLMs in motion understanding and generation remains unexplored.
\textit{(3) Body-only motion representations}. 
Existing text-based motion generation  solutions~\cite{zhang2023t2m,zhang2022motiondiffuse,zhang2023remodiffuse,guo2022generating,guo2024momask} primarily focus on generating body-only motions rather than holistic motions. However, their plausibility and expressiveness remain unsatisfactory in certain scenarios (\textit{e.g.}, sports activities and playing musical instruments), as important details of human motion—\textit{e.g.}, fine-grained hand gestures, are frequently overlooked.
% Existing solutions~\cite{zhang2023t2m,zhang2022motiondiffuse,guo2022generating} focus mainly on body motion, overlooking fine-grained hand gestures, limiting expressiveness in scenarios like musical instrument playing.

% Recently, Motion-LLM~\cite{wu2024motionllm} provides a streamlined approach for unleashing the potential of pre-trained LLMs in motion-related tasks. Yet, despite its effectiveness, its lack of extension to 3D whole-body motions diminishes the expressiveness and physical plausibility of the human motions.
% Therefore, it is important to develop a unified human motion comprehension and generation framework that can efficiently utilize multiple control signals simultaneously. \textit{Body-only motion representations}. 

% Motivated by these limitations, our objective is to develop a Large Motion-Language Model (LMLM), capable of handling \textbf{diverse control signal modalities}, \textbf{various motion-related tasks}, and \textbf{holistic motion representation}. As outlined above, three challenges are crucial and need to be solved: 
To overcome these limitations, we propose MotionGPT-2, a unified Large Motion-Language Model (LMLM) capable of handling diverse control signals, performing various motion-relevant tasks, and generating holistic human motions. 
% Our proposed framework builds on previous work, extending MotionGPT~\cite{jiang2023motiongpt} with new capabilities. 
The key innovations of MotionGPT-2 include:
(1) \textbf{Formulating Multimodal Control Signals into a Unified Representation}: 
% In previous work, MotionGPT~\cite{zhang2023motiongpt}, we design a versatile framework and task-aware prompts for human motion synthesis. 
MotionGPT-2 designs a versatile framework and task-aware prompts for human motion synthesis. This framework, in particular, allows for the generation of human motions governed by multimodal control conditions, described by ${M_{\text{out}}, T_{\text{out}}} = f(T_{\text{in}}, \text{task}, M_{\text{in}})$. Here $M_{\text{in}}$ and $T_{\text{in}}$ refer to the input motions and texts, while $M_{\text{out}}$ and $T_{\text{out}}$ represent the resulting outputs. The variable $\text{task}$ indicates the task-aware prompts that adapt the model to specific motion-related tasks. 
Compared to MotionGPT~\cite{zhang2023motiongpt}, MotionGPT-2 broadens the scope by incorporating additional motion-related tasks, \textit{i.e.}, motion captioning, general motion completion. 
% For example, in the motion prediction task, when provided with $M_{\text{in}}$ and tasked with ``Generate motion given initial poses," the model predicts subsequent frames based on the initial poses. 
% This versatile framework offers a comprehensive solution for human motion synthesis, where task instructions and multimodal conditions contribute to generating more controllable motions.

% (2) \textit{How to build a task-agnostic framework and achieve strong generalization across various motion-related tasks}? Our conference version MotionGPT~\cite{zhang2023motiongpt} has argued that the challenge is naturally resolved assisted by LLMs. There are several compelling reasons. First, recent studies have demonstrated that LLMs can understand multimodal inputs, \textit{e.g.,} images~\cite{zhu2023minigpt,du2023vision,li2023blip, liu2023llava, ye2023mplugowl} and videos~\cite{2023videochat}, through a lightweight adapter~\cite{hu2021lora}. Therefore, we expect the LLMs can also understand motion sequences with an appropriate adapter. Second, LLMs can provide diverse human motion contexts for motion generation because they have encoded diverse motion patterns from extensive large-scale text data. This enables our motion generator fine-tuned from LLMs can produce motions with rich patterns. Third, since LLMs output tokens aggressively, producing human motion with flexible sequences is no longer an obstacle. 
(2) \textbf{Building a Task-Agnostic Framework with Strong Generalization}: 
Our conference version, MotionGPT~\cite{zhang2023motiongpt}, argued that this challenge can be naturally addressed with the assistance of LLMs. There are several compelling reasons. First, recent investigations indicate that LLMs have the ability to comprehend inputs from multiple modalities (\textit{e.g.}, images~\cite{zhu2023minigpt,du2023vision,liu2023llava,ye2023mplugowl} and videos~\cite{2023videochat}) through lightweight adapters~\cite{hu2021lora}. Therefore, we expect that with suitable adapters, LLMs will be capable of understanding motion sequences. Second, LLMs have learned a broad range of motion patterns from their extensive text training, which allows them to offer diverse human motion contexts for generating motion. Consequently, our motion generator, adapted from LLMs, can produce motions with a wide range of rich patterns. Third, because LLMs generate tokens in a sequential manner, generating human motion with adaptable sequences is now easily achievable.
% \textcolor[RGB]{255,153,51}{Although the previous MotionGPT~\cite{zhang2023motiongpt} has achieved significant success in developing a unified human motion generation framework, it has not fully leveraged the complex relationships between motion and the contextual cues present in nonverbal communication. In fact, the nonverbal motion sequences carry rich information, \textit{e.g.}, intentions and environmental interactions, similar to contextual semantic in language. }

% The nonverbal motion sequences carry rich information, \textit{e.g.}, intentions and environmental interactions, acted as a specialized ``body language''. 

In this work, we introduce MotionGPT-2, an enhanced version of MotionGPT~\cite{zhang2023motiongpt} by using LLMs to jointly represent motion and language. We first embed the human motions into discrete motion tokens via the Vector Quantized Variational-AutoEncoder (VQ-VAE), which efficiently captures the underlying structure of human motion. Then, we expand the LLM’s vocabulary with these motion tokens, creating an enriched motion-language vocabulary. By incorporating human motion and language into a unified vocabulary, our model fosters a more transparent understanding of the complex relationships between the two.  Further, MotionGPT-2 integrates tokens from both language and motion prompts to generate instructions. We implement a multimodal pre-training combined with an instruction-tuning approach to train MotionGPT-2 efficiently, utilizing the established LoRA adaptation method. The motion instruction tuning framework we have developed allows for the integration of pose sequence information into the fine-tuned LLM, while capitalizing on the strong motion priors inherent in the origin LLM. With mere 1\% parameters, the generalized MotionGPT-2 achieves competitive results in multiple motion-related tasks compared to those trained-from-scratch models with specialized frameworks. 

(3) \textbf{Achieving Precise Discrete Representations of Holistic Human Motions}: 
To address this issue, we incorporate kinematic structure priors and design an innovative Part-Aware VQ-VAE for holistic motion tokenization. Compared to the vanilla motion VQ-VAE used in MotionGPT~\cite{zhang2023motiongpt}, the proposed Part-Aware VQ-VAE utilizes two-level discrete codebooks and motion encoders to learn body-hand representations. The key insight of our Part-Aware VQ-VAE lies in its ability to learn informative and compact representations of fine-grained holistic motions. This two-level discretization framework captures subtle hand movements while maintaining global body dynamics. Finally, the hand and body motion tokens are integrated into the LLM's vocabulary, enabling MotionGPT-2 to correctly interpret and generate holistic motion-related sequences in response to instructions.

% To capture fine-grained motion details, we introduce Part-Aware VQ-VAE, an enhanced discretization framework that models both body and hand movements. Compared to the vanilla VQ-VAE in MotionGPT~\cite{zhang2023motiongpt}, the Part-Aware VQ-VAE uses two-level discrete codebooks to represent global body dynamics and subtle hand gestures. These motion tokens are integrated into the LLM’s vocabulary, enabling precise interpretation and generation of holistic motion sequences.

% We conduct extensive experiments on the HumanML3D~\cite{guo2022generating}, KIT-ML~\cite{plappert2016kit} datasets, demonstrating MotionGPT-2 has a strong ability for multiple motion-related tasks by fine-tuning an LLM.
% Notably, MotionGPT achieves this with a minimal set of training parameters, and in less training time (just 10\% of the time taken by other methods). We also observe that joint training under multiple control instructions surpasses training with singular control signals, showing the effectiveness of our unified motion-language fine-tuning paradigm. Experiments on Motion-X~\cite{lin2024motion} dataset verifies that our proposed MotionGPT-2 is highly adaptable to the challenging 3D holistic motion generation task. This paper is an extended version of our conference paper~\cite{zhang2023motiongpt}, where we make several \textbf{new contributions}: 
We conduct extensive experiments on the HumanML3D~\cite{guo2022generating}, KIT-ML~\cite{plappert2016kit} dataset, demonstrating that MotionGPT-2 has strong adaptability and efficiency across multiple motion-related tasks by fine-tuning an LLM. With only 1\% of additional parameters, MotionGPT-2 achieves highly competitive performance across tasks while significantly reducing training time—requiring only 10\% of the time compared to other methods. We also observe that joint training under multiple control instructions surpasses training with singular control signals, highlighting the effectiveness of the unified motion-language fine-tuning paradigm. Experiments on the Motion-X~\cite{lin2024motion} benchmark verify that the MotionGPT-2 is adaptable to the challenging 3D holistic motion generation task. MotionGPT-2 sets a new benchmark for motion-related foundational models.

In summary, we extend our conference version~\cite{zhang2023motiongpt} \textbf{by making several additional novel contributions}:

\begin{itemize} 

\item We propose an enhanced version of the previous MotionGPT, MotionGPT-2. Compared to its predecessor, MotionGPT-2 serves as a unified Large Motion-Language Model to handle multiple motion-related tasks, enabling us to establish new state-of-the-art performance benchmarks.

\item By creating an enriched motion-language vocabulary, we empower the pre-trained LLMs with the ability to unify the understanding and generation of body kinetics.
 
\item We further extend the proposed MotionGPT-2 to tackle the challenging 3D whole-body motion generation task by introducing a whole-body motion discretization framework, \textbf{Part-Aware VQ-VAE}, which encodes body motions and hand gestures with two structured codebooks for fine-grained motion representations.
\end{itemize}

We conduct extensive experiments on the HumanML3D, KIT-ML, and Motion-X datasets to validate the superiority of our proposed LLM-based unified motion-language model across multiple motion-related tasks. We provide an in-depth analysis and visualizations of MotionGPT-2, indicating that further advancements in LLM technology hold the potential to enhance its performance in the future. 

\section{Related Work}
\subsection{LLMs and Multi-modal Large Language Models}
Large Language Models (LLMs)~\cite{devlin2018bert,radford2018improving,radford2019language,brown2020language,OpenAI2023GPT4TR,touvron2023llama} have experienced rapid development recently, \textit{e.g.,} BERT~\cite{devlin2018bert}, GPT~\cite{radford2018improving}, and Google T5~\cite{raffel2020exploring}.
Notably, models like GPT-4~\cite{OpenAI2023GPT4TR} show outstanding performance on various language tasks due to their extensive training datasets (GPT-4 utilizes 45 gigabytes) and numerous parameters.
Traditionally, language models were crafted for specific tasks like translation or sentiment analysis, but recent innovations, such as ChatGPT, have broadened their functional scope. Built on the GPT-4 framework, ChatGPT can engage in interactive dialogues, showcasing strong natural language understanding. This effectiveness opens new avenues for downstream applications achievable through the fine-tuning of these LLMs. Nevertheless, fine-tuning these extensive models poses significant challenges. To mitigate this issue, several efficient fine-tuning techniques have emerged, including prompt tuning~\cite{lester2021power,liu2021p,hu2021knowledgeable}, adapters~\cite{houlsby2019parameter,he2021effectiveness,le2021lightweight}, and LoRA~\cite{hu2021lora}. Our study is inspired by recent developments in LLMs while tackling a different problem through the integration of a novel modality.

% \begin{figure*}[t]
%     \centering
%     \includegraphics[width=\textwidth]{pipeline.pdf}
%     \caption{
%     The pipeline of MotionGPT, a Motion General-Purpose generaTor.
%     Given text and poses as an input example, we organize task descriptions (Instruction) and multiple control conditions (Input) within a question template. 
%     MotionGPT fine-tunes an LLM to generate the corresponding motion answer, which can then be decoded into human motions using a VQ-VAE decoder.
%     }
%     \label{fig:pipeline}
% \end{figure*}

\vspace{-0.37cm}
\subsection{Human Motion Generation}
Motion generation has a long-standing history in research~\cite{tevet2022motionclip,habibie2017recurrent,li2017auto,zhang2022motiondiffuse,guo2020action2motion,tevet2022human,petrovich2022temos,li2021ai} and can be conditioned on various inputs, including motion descriptions, specific actions, and music.
For instance, HP-GAN~\cite{barsoum2018hp} and \cite{martinez2017human} adopt a sequence-to-sequence model to forecast future poses from earlier ones. 
Moreover, ACTOR~\cite{petrovich2021action} utilizes a transformer-based VAE for both unconditional generation and action-driven motion generation.
TRAJEVAE~\cite{kania2021trajevae} generates motion sequences that align with a given trajectory when provided with an initial pose. 
Recently, a significant focus has been placed on text-conditional motion generation, which involves creating human motion sequences based on textual prompts.Previous researches~\cite{guo2022generating,ghosh2021synthesis,ahuja2019language2pose,tevet2022motionclip,zhang2023t2m} focus on modeling a joint latent space for motion and text alignment.
In TEMOS~\cite{petrovich2022temos}, a VAE model is proposed that creates a shared latent space for motion and text interactions. T2M-GPT~\cite{zhang2023t2m} formulates the motion generation as the next index prediction task and leverages small language models to model the translation mapping between discrete motion indices and text. MotionCLIP~\cite{tevet2022motionclip} and TM2T~\cite{guo2022tm2t} align the shared space of text and motion with the expressive CLIP~\cite{radford2021learning} embedding space. MotionDiffuse~\cite{zhang2022motiondiffuse} introduces a diffusion model into its framework for generating motion from textual descriptions, which leads to impressive outcomes. In a different approach, MDM~\cite{tevet2022human} aims to boost the coherence between motion and textual inputs by implementing CLIP~\cite{radford2021learning} as the text encoder, thereby enhancing the model with more effective textual priors. However, it is challenging to model semantically complex relationships of human motions and texts, particularly in the absence of general world knowledge, \textit{e.g.}, \textit{how specific gestures convey intentions} and \textit{how to interpret body language in different context}. Current efforts, such as MotionGPT~\cite{jiang2023motiongpt}, MotionLLM~\cite{wu2024motionllm}, M3-GPT~\cite{luo2024m} have initiated the development of a unified motion-language model aimed at generating plausible human motions along with textual descriptions driven by prompt instructions. Although remarkable achievements have been made in multiple motion-related tasks, these approaches, which only utilize relatively small language models, overlook exploration of the potential of LLMs in understanding and generating human motions. Moreover, they are inadequate in addressing the complex body-hand representations. Unlike prior approaches, MotionGPT-2 is distinguished as the first Large Motion-Language Model (LMLM) capable of multimodal control, handling diverse motion-related tasks, and providing a comprehensive representation of motion.
\vspace{-0.3cm}
% \subsection{Overview}

% By asking the Large Language Model (LLM) to generate desirable human motions according to task prompts and control conditions.
% \textcolor[RGB]{255,153,51}{Following the typical framework of MLLM~\cite{liu2023llava,lai2024lisa}, we leverage cross-entropy loss to supervise the LoRA adapter.}
\begin{figure*}[t]
    \centering
    \includegraphics[width=\textwidth]{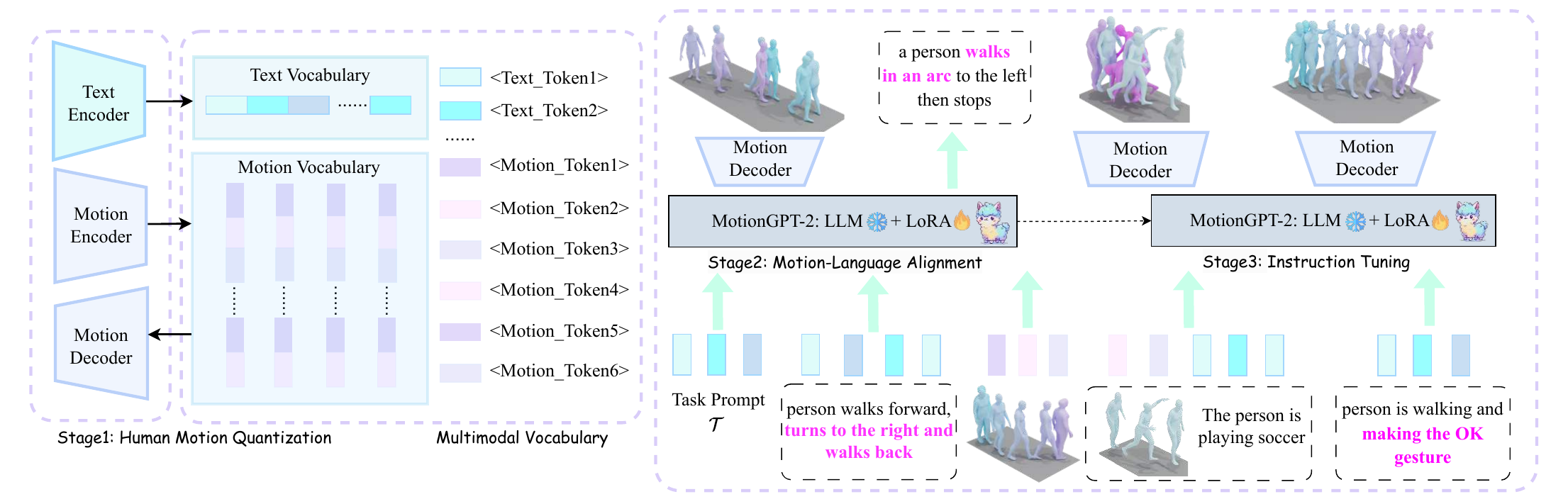}
    \caption{
    MotionGPT-2 consists of multi-modal tokenizers (Section~\ref{3D Human Motion Quantization}) and a motion-language model (Section~\ref{sec: Muiti-modal Motion-Language Learning}). With unified multimodal vocabulary and task-aware instructions (Section~\ref{sec: instruction generation}). It enables to accept multiple control conditions (Section~\ref{sec: generalization to existing and new tasks}). MotionGPT-2 is learned by the \textit{Motion-Language Alignment} stage and \textit{Instruction Tuning} stage (Section~\ref{Training Strategy}).
    }
    \label{fig:pipeline-MotionGPT2}
    \vspace{-0.45cm}
\end{figure*}

\section{Motion General-Purpose generaTor (MotionGPT)}

Our conference version MotionGPT acts as a Motion General-Purpose generator that is driven by multi-modal inputs. To align with the LLM’s token-in-token-out nature, we first quantize the human motions into code representations via the well-established motion VQ-VAE~\cite{van2017neural} (Section~\ref{3D Human Motion Quantization}). These motion discrete codes, along with text control conditions and specially-crafted task-aware instructions, are organized into a unified question template (Section~\ref{sec: instruction generation}). By adjusting the task-aware instructions, MotionGPT reveals its potential as a generic baseline framework for tasks involving motion generation.

\vspace{-0.4cm}
\subsection{3D Human Motion Quantization}
\label{3D Human Motion Quantization}
VQ-VAE proposed in \cite{van2017neural} is designed to learn discrete representations with semantic-rich codebooks in an encoding-decoding fashion. To discretely represent human motions as semantic tokens, we introduce the motion VQVAE model, which consists of a motion encoder $\mathcal{E}$, a motion decoder $\mathcal{D}$, and a codebook $\mathcal{B}_m=\{b_1, b_2, \dots, b_N\}$, where $N$ denotes the codebook size. We denote a human motion as $\mathbf{m}\!=\!\{m_1,m_2,...,m_T\}\!\in\!\mathbb{R}^{T\times d}$, where $T$ is the motion sequence length, and $d$ is the dimension per frame.

To learn the codebook of human motions, the estimated motion embedding $\mathcal{E}(\mathbf{m})$ is transformed into a collection of codebook entries through quantization. Further, the quantized motion vector is obtained by searching the nearest corresponding embedding in a learnable codebook $\mathcal{B}_m$, which can be mathematically formulated as: 
\begin{equation}
\label{eq: vq-vae-0}
    \mathbf{e} = \mathop{\arg\min}\limits_{b_k\in\mathcal B}\|\mathcal{E}(\mathbf{m})-b_k\|_2.
\end{equation}

Based on the estimation latent representation $\mathbf{e}$, the motion decoder $\mathcal{D}$ employs several 1-D deconvolutional layers to extract frame-wise motion features, decoding the sequential codes $\mathbf{e}$ to the raw motion space as the motion $\mathbf{\hat{m}}$: $\mathbf{\hat{m}} = \mathcal{D}(\mathbf{e})$. The motion code $p$ of the human pose $\mathbf{m}$ can be calculated as the index of its nearest embedding in the codebook, \textit{i.e.,}
\begin{equation}
    \quad p = \mathop{\arg\min}\limits_{k}\|\mathcal{E}(\mathbf{m})-b_k\|_2.
    \label{eq: vq-vae}
\end{equation}

% Based on the estimated latent representation $\mathbf{e}$ of the motion $\mathbf{m}$, the reconstructed human pose $\mathbf{\hat{m}}$ can be produced by the decoder of VQ-VAE and the motion code $p$ of human pose $\mathbf{m}$ can be calculated as the index of its nearest embedding in the codebook, \textit{i.e.,} 
% \begin{equation}
%      \mathbf{\hat{m}} = \mathcal{D}(\mathbf{e}), %, \quad p = \mathop{\arg\min}\limits_{k}\|\mathcal{E}(\mathbf{m})-b_k\|_2.
%     \label{eq: vq-vae}
% \end{equation}

The standard motion VQ-VAE can be trained by the three tailored loss: 1) a reconstruction loss to minimize the distance between the decoded motion $\mathbf{\hat{m}}$ and the origin motion $\mathbf{m}$, 2) a codebook loss to encourage the $b_k$ to be drawn closer to the encoded embedding $\mathcal{E}(\mathbf{m})$, aligning the discrete code representation with origin motion embeddings. 3) a commitment loss to guide the encoded embedding $\mathbf{e}$ to remain close to the corresponding discrete code $b_k$, stabilizing the training process and reducing codebook oscillation. The total loss function $\mathcal{L}_{\text{VQVAE}}$ can be formulated as follows:
\begin{equation}
\begin{aligned}
    \mathcal{L}_{\text{VQVAE}} = ||\mathcal{D}(\mathcal{E}(\mathbf{m}))-\mathbf{m}||^2 + \|\operatorname{sg}[\mathcal{E}(\mathbf{m})] - \mathbf{e}\|^2_2 \\
    + \beta\|\mathcal{E}(\mathbf{m}) - \text{sg}[\mathbf{e}]\|^2_2.
    \label{vqvae_loss}
\end{aligned}
\end{equation}

\noindent Here, $\operatorname{sg}$ indicates the stop gradient operation and $\beta$ is the hyper-parameter to control the weight of the commitment loss. To enhance the quality of the generated motion, we incorporate \textit{L1 smooth loss} and \textit{velocity regularization loss} in the reconstruction loss. The codebook is optimized with Exponential Moving Average (EMA) operation and codebook reset techniques following \cite{jiang2023motiongpt,zhang2023t2m,zhang2023motiongpt}.

\vspace{-0.45cm}
\subsection{Instruction Generation}
\label{sec: instruction generation}
In our previous conference version, we design instructions that integrate task prompts and control conditions to enable (text, motion)-motion generation tasks. Given the task prompts $\mathcal{T}\!=\!\{t_1, t_2, ..., t_{n_t}\}$, the text control conditions $\mathcal{X}=\{x_1, x_2, ..., x_{n_x}\}$ and the pose control conditions $\mathcal{P}=\{p_1, p_2, ..., p_{n_p}\}$ where $n_t$, $n_x$ and $n_p$ are the number of codes in $\mathcal{T}$, $\mathcal{X}$ and $\mathcal{P}$, the instruction template $\mathcal{I}$ is formulated as:
\vspace{0.2cm}
\begin{mdframed}[backgroundcolor=gray!20]
\textcolor{blue}{\% General control conditions format} \\
Control Conditions: \{The Text control condition $\mathcal{X}$ $<$$x_1, x_2, ..., x_{n_x}$$>$\} \{The Pose control condition $\mathcal{P}$ $<$$p_1, p_2, ..., p_{n_p}$$>$\} \\
\textcolor{blue}{\% General instruction format } \\
\textbf{Instruction} ($\mathcal{I}$):
\{The Task Prompt $\mathcal{T}$ $<$$t_1, t_2, ..., t_{n_t}$$>$\} 
\{Control Conditions\}
\end{mdframed}
\vspace{0.2cm}

Pose control conditions $\mathcal{P}=\{p_1, p_2, ..., p_{n_p}\}$, representing pose codes produced by the previously mentioned motion VQ-VAE. The entire instruction $\mathcal{I}$ is conceptualized as a series of specialized text inputs.
By formulating diverse instruction prompts, the MotionGPT~\cite{zhang2023motiongpt} tackles both traditional motion generation tasks and emerging challenges in motion synthesis.

Specifically, for text-based motion generation task, MotionGPT address it by instantiating following instruction $\mathcal{I}$:
\begin{mdframed}[backgroundcolor=gray!20]
\textbf{Instruction} ($\mathcal{I}$): \{\textcolor{blue}{\textbf{Task Prompts:}} ``Generate a sequence of motion tokens matching the following human motion description.''\} \{\textcolor{blue}{\textbf{Control Conditions:}} Text control condition $\mathcal{X}$\}
\end{mdframed}
% \vspace{0.2cm}

\noindent By adjusting instructions, MotionGPT is seamlessly adapted to a wide variety of control conditions, \textit{e.g.} the textual description and an arbitrary number of human poses, making it a highly versatile and flexible solution for motion-related tasks.
% % \vspace{0.2cm}
% \begin{mdframed}[backgroundcolor=gray!20]
% \textbf{Instruction} ($\mathcal{I}$): \{\textcolor{blue}{\textbf{Task Prompts:}} ``Generate a sequence of motion tokens matching the following human motion description given the init/last/key pose tokens."\} \{\textcolor{blue}{\textbf{Control Conditions:}} Text control condition $\mathcal{X}$ $<$Motion Token$>$ Pose control conditions $\mathcal{P}$ $<$/Motion Token$>$\}
% \end{mdframed}
% % \vspace{0.2cm}

% For the motion generation task, the answer of LLM is represented as $\mathcal{\hat{P}}= \{\hat{p}_1, \hat{p}_2, ..., \hat{p}_{n_{\hat{p}}}\}$, comprising a sequence of generated motion codes. These codes can be decoded to human motion using Eq.~\ref{eq: vq-vae}.

\vspace{-0.25cm}
\subsection{Model Optimization}
In the previous MotionGPT~\cite{zhang2023motiongpt}, we utilize a decoder-only LLM following~\cite{wu2024motionllm}, which effectively handles complex relationships and underlying patterns between text and motion. The token-in-token-out LLM maximizes the probability $p_\theta\left(x_t \mid x_{<t}, \mathcal{T}, c\right)$ of succeeding token in an autoregressive manner. Here, $\mathcal{T}$ represents the Task Prompts, $c$ is the control signal, $x_{1:T}$ denotes the target token sequence. Therefore, during the training process, the Cross Entropy loss is applied to ensure the correspondence between the estimated tokens and the real tokens, fine-tuning the LLM by LoRA~\cite{hu2021lora}, which is mathematically formalized as:
\begin{equation}
\mathcal{L}_{\text{LoRA}} = -\sum \log p_\theta\left(x_t \mid x_{x_{<t}}, \mathcal{T}, c\right).
\end{equation}

\begin{figure*}[t]
    \centering
    \includegraphics[width=\textwidth]{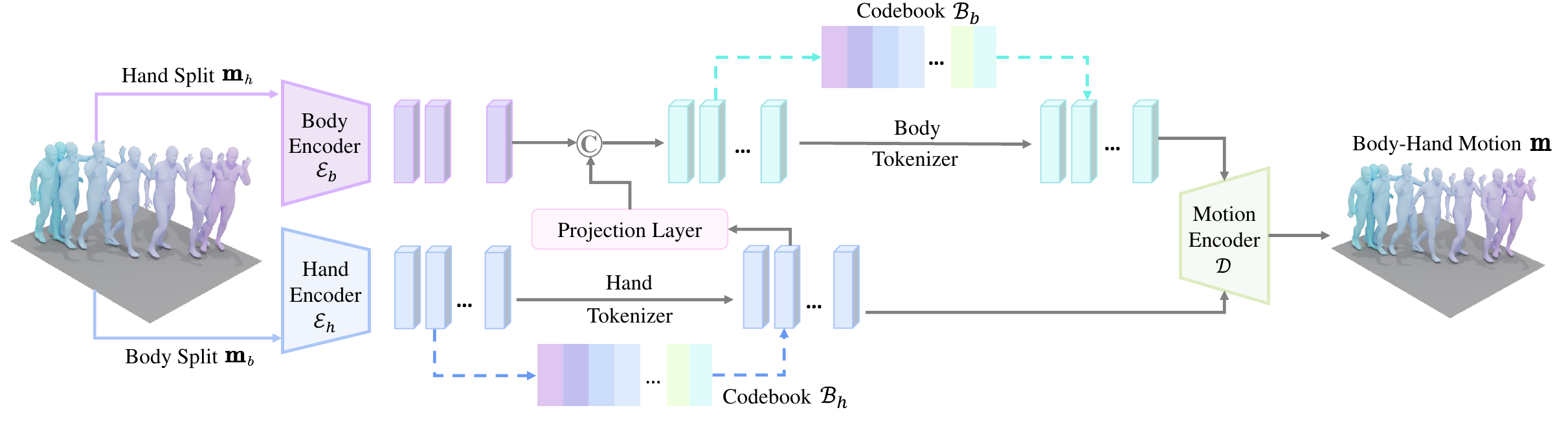}
    \caption{
   The framework overview of our proposed Part-Aware VQVAE for body-hand motion tokenization. The Part-Aware VQVAE splits SMPL-X-based human representations into body-hand motions $\bf{m}_b$ and $\bf{m}_h$. Then, it quantizes fine-grained body-hand motion into two discrete codebooks $\mathcal{B}_b$ and $\mathcal{B}_h$ with hierarchical body priors.
    }
    \label{fig:Part-Aware-VQVAE}
    \vspace{-0.5cm}
\end{figure*}

\vspace{-0.3cm}
\section{MotionGPT-2: A Versatile Model Supporting Holistic Motion Understanding and Generation}

Building upon the MotionGPT, MotionGPT-2 (as shown in Figure.~\ref{fig:pipeline-MotionGPT2}) is to formulate a \textbf{multi-task unified} (motion generation, motion captioning, and generalized motion completion) framework to support \textbf{holistic motion representations} (fine-grained body and hand movements). 
The key designs are as follows: (1) \textit{Part-Aware VQ-VAE for Body-hand Tokenization.} By designing an innovative Part-Aware VQVAE, our MotionGPT-2 achieves informative and compact representations of fine-grained holistic motions, making it adaptable to holistic motion generation tasks (Section~\ref{sec: MotionGPT2 for Holistic Human Motions}). (2) \textit{A Unified Motion-Language Vocabulary}. By expanding the text vocabulary of LLM, we develop a unified Large Motion-Language Model (LMLM) (Section~\ref{sec: Muiti-modal Motion-Language Learning}) for the seamless integration of motion and language modalities. This vocabulary extension enables MotionGPT-2 to support a broader range of novel motion-related tasks by different instructions (Section~\ref{sec: generalization to existing and new tasks}). (3) \textit{Three-Stage Training Strategy}. To enhance alignment between motions and texts, we develop a three-stage training pipeline (Section~\ref{Training Strategy}), including Motion Tokenization, Motion-Language Alignment, and Instruction Fine-tuning.
% \textcolor[RGB]{255,153,51}{By simply adjusting task-aware instructions, we demonstrate the potential of MotionGPT-2 as \textbf{a generic baseline framework} for motion-related tasks.}
\vspace{-0.5cm}

\subsection{Part-Aware VQVAE for Holistic Motion Representations}
\label{sec: MotionGPT2 for Holistic Human Motions}
Our previous MotionGPT is limited to the generation of body-only motions. Yet, advancing toward holistic motion generation is essential for achieving more realistic and lifelike animations. Holistic motions are considerably more complex than their body-only counterparts, necessitating the detailed and coordinated motion generation involving the body and hand features. In this section, our MotionGPT-2 emphasizes the representation of body-hand motions. With the above single motion vocabulary approach, the embedding of holistic human motion inevitably leads to ambiguities where similar actions are represented by the same motion token. To model distinct semantic granularity for body and hand, we design a Part-Aware VQVAE module, which utilizes the codebook $\mathcal{B}_b$ and $\mathcal{B}_h$ to learn discrete representations for body and hand, in a more structured and coordinated manner. 

As illustrated in Fig.~\ref{fig:Part-Aware-VQVAE}, we input the SMPL-X based body-hand representations $\mathbf{m}^{B}\!=\!\{m^{B}_1,m^{B}_2,...,m^{B}_T\}\!\in\!\mathbb{R}^{T\times d}$ and $\mathbf{m}^{H}\!=\!\{m^{H}_1,m^{H}_2,...,m^{H}_T\}\!\in\!\mathbb{R}^{T\times d}$ into two separate encoders. As shown in Fig.~\ref{fig:Part-Aware-VQVAE}, the hand embedding $\mathbf{e}_h$ is first quantized by the hand codebook $\mathcal{B}_h$ and we then fuse the body and hand tokens via the concatenation operator before quantizing the body embedding $\mathbf{e}_b$ using the body codebook $\mathcal{B}_b$. The interaction of body and hand motions is tightly coupled and articulated, shaped by biomechanical and physical restrictions in the context of complex activities. Such a \textit{fuse-before-quantize} method for body tokens effectively enhances the overall naturalness and coordination of holistic motion representations. Akin to the \cite{lu2023humantomato}, the Body-hand Decoder is designed to take body tokens $p^B$ and hand tokens $p^H$ as input to accurately reconstruct corresponding motions.

\vspace{-0.3cm}
\subsection{A Unified Motion-Language Vocabulary}
\label{sec: Muiti-modal Motion-Language Learning}

Most previous motion-related studies~\cite{zhang2023t2m,zhang2023motiongpt,zhang2024motiondiffuse,zhang2023remodiffuse,guo2022tm2t} have viewed textual descriptions and motions as separate modalities. However, similar to sentences in natural language, a continuous motion $\mathbf{m}$ is compressed into motion tokens, serving as a type of ``body language''. Therefore, our MotionGPT-2 naturally uses LLMs to jointly model motion and language representations. It reflects how humans execute body motions, allowing for seamless transitions across motion-related modalities in real-life scenarios.

Specifically, to merge multi-modal discrete tokens with the pre-trained LLM, the original vocabulary $\mathcal{B}_t$ is expanded to incorporate motion tokens $\mathcal{B}_m$. Further, we also insert several special tokens $\mathcal{B}_s$ into the LLM's vocabulary, \textit{e.g.}, $\left<\text{motion}\right>$ and $\left</\text{motion}\right>$, which signify the beginning and end of the motion. This extends the vocabulary into a unified text-motion set, $\mathcal{B} = \{\mathcal{B}_t, \mathcal{B}_m, \mathcal{B}_s\}$. The newly incorporated parameters are initialized randomly. Expanding the LLM’s vocabulary, a human motion is denoted as a token sequence that is LLM-understandable. Equipped with the vocabulary $\mathcal{B}$, we model the joint distribution of textual descriptions and human motions in a unified space. It allows us to formulate motion-related tasks in a general framework, leveraging the task-aware instructions.

\vspace{-0.3cm}
\subsection{Instruction Prompts in New Motion-Related Tasks}
\label{sec: generalization to existing and new tasks}
Our previous MotionGPT primarily focused on motion generation tasks. In contrast, our MotionGPT-2 enhances its capabilities by defining several core motion-related tasks, including motion captioning, motion prediction, and motion interpolation. For each task, we construct unique instruction prompts tailored to the specific task requirements. 
% With the general template given before, our MotionGPT-2 is capable of being a general-purpose motion generator and captioner, supporting various motion-related tasks.
% Specifically, for existing text-based motion generation setting, MotionGPT-2 address it by constructing following instruction $\mathcal{I}$:
% \begin{mdframed}[backgroundcolor=gray!20]
% \textbf{Instruction ($\mathcal{I}):$} \{\textcolor{blue}{\textbf{Task Prompts:}} ``Generate a sequence of motion tokens matching the following human motion description."\} \{\textcolor{blue}{\textbf{Control Conditions}:} Text control condition $\mathcal{X}$\}
% \end{mdframed}
For instance, given a LLM $\mathcal{F}$, the instruction template $\mathcal{I}$ for motion captioning and motion prediction task as well as the answer of the LLM $\mathcal{\hat{P}}=\mathcal{F}(\mathcal{I})$ are defined as: 
\begin{mdframed}[backgroundcolor=gray!20]
Below is an instruction that describes a task, paired with an input that provides further context. Write a response that appropriately completes the request. \\
\textcolor{blue}{\% \textbf{Task Prompts}: Motion Captioning Task Prompts} \\
\textcolor{blue}{\% \textbf{Control Conditions}: Code Sequences of Control Conditions (Motion Tokens)} \\
% \textcolor{gray}{\% Control Conditions: Code sequences of Control Conditions} \\
\textbf{Instruction $\mathcal{I}$:} \{Task Prompts $\mathcal{T}$\}\{Control Conditions\} \\
\textbf{Answer $\mathcal{\hat{P}}$: } \{Sequences of Text Tokens\} \\
\textcolor{blue}{\% \textbf{Task Prompts}: Motion Prediction Task Prompts} \\
\textcolor{blue}{\% \textbf{Control Conditions}: Code Sequences of Control Conditions (Initial Several Motion Tokens)} \\
\textbf{Instruction $\mathcal{I}$:} \{Task Prompts $\mathcal{T}$\} \{Control Conditions\}\\
\textbf{Answer $\mathcal{\hat{P}}$: } \{Sequences of Human Motion Tokens\} 
\end{mdframed}
\vspace{-0.1cm}

In the text-based holistic motion generation task, the instruction tuning phase proceeds after obtaining the discrete motion tokens $p^B$ and $p^H$. During this stage, we formulate the instruction template $\mathcal{I}$ and the answer $\mathcal{P}$ of the LLM for motion generation. Inspired by MLLMs~\cite{zhu2023minigpt,liu2023llava}, we unify the discrete body-hand tokens within a general prompt template.
% \vspace{0.2cm}

\begin{figure*}[t]
    \centering
    \includegraphics[width=1\textwidth]{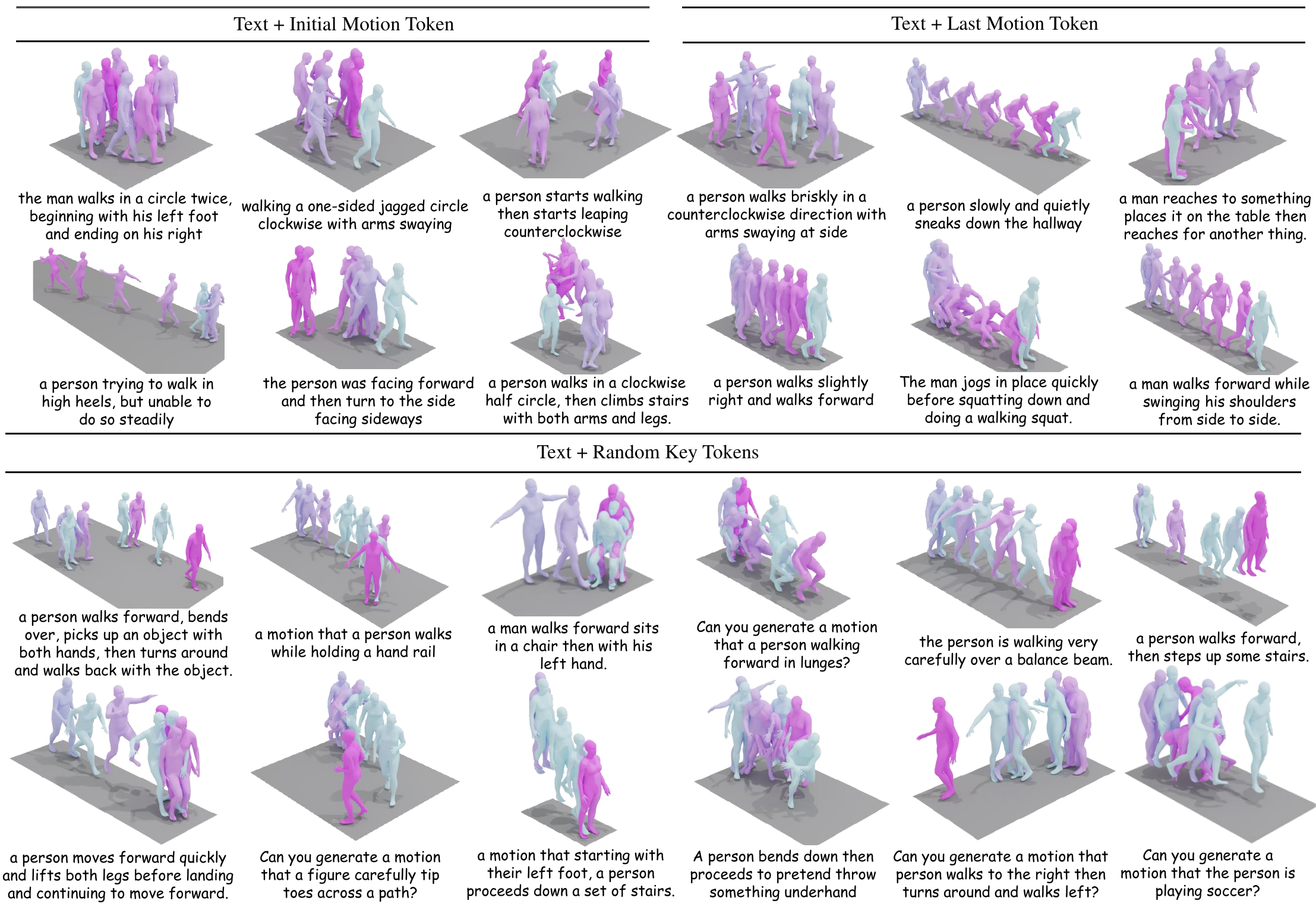}
    \caption{
    Gallery showcasing the results of generated human motions by MotionGPT-2 with multiple control conditions on the HumanML3D dataset~\cite{guo2022generating}, \textit{i.e.}, \textit{Text+Initial Motion Token}, \textit{Text+Last Motion Token}, and \textit{Text+Random Key Motion Token}. With these diverse control signals, our MotionGPT-2 demonstrates the ability to generate physically realistic human motions.
    }
    \label{fig:multiple-control}
    \vspace{-0.5cm}
\end{figure*}

\begin{mdframed}[backgroundcolor=gray!20, innerleftmargin=3pt, innerrightmargin=3pt, innerbottommargin=0pt]
Below is an instruction that describes a task, paired with an input that provides further context. Write a response that appropriately completes the request. \\
\textcolor{blue}{\% \textbf{Task Prompts}: Motion Generation Task Prompts $\mathcal{T}$} \\
\textcolor{blue}{\% \textbf{Control Conditions}: Sequences of Control Conditions (Text Tokens)} \\
\textbf{Instruction $\mathcal{I}$:} \{Task Prompts $\mathcal{T}$\} \{Control Conditions\}\\
\textbf{Answer $\mathcal{\hat{P}}$: } \{$<{\text{Motion}}\_{\text{Body}}\_{\text{Token}}><{\text{Body}}\_{\text{Token}}></\text{Motion}\_{\text{Body}}\_{\text{Token}}><\text{Motion}\_{\text{Hand}}\_{\text{Token}}><\text{Hand}\_{\text{Token}}></\text{Motion}\_{\text{Body}}\_{\text{Token}}>$\} 
\end{mdframed}
\vspace{-0.45cm}

\subsection{Three-Stage Training Strategy}
\label{Training Strategy}

Our previous MotionGPT~\cite{zhang2023motiongpt} is founded on a two-stage approach. Nevertheless, the absence of motion-language alignment hinders model from interpreting the complex relationships between motions and texts. Towards this issue, MotionGPT-2 proposes a comprehensive three-stage training strategy:

\textbf{Stage 1: Training of Motion Tokenizer.} As depicted in Section~\ref{3D Human Motion Quantization}, we first learn the discrete motion representations to align with the LLM’s token-in-token-out nature. Guided by the objective outlined in Eq.~\ref{vqvae_loss}, the quantization process allows human motions $\mathbf{m}$ to be expressed as a sequence of tokens, which seamlessly integrates with descriptive text. To maintain stability in LLM optimizing, we subsequently freeze weights of motion-aware VQVAE during further training stages.

\textbf{Stage 2: Motion-Language Alignment.} Inspired by current Multi-modal Large Language Models (MLLMs)~\cite{liu2023llava,zhu2023minigpt}, to align the motion and language feature space of MotionGPT-2, the LLaMA~\cite{touvron2023llama} model is fine-tuned on a wide range of motion-related tasks. \textit{To accurately interpret contextual `body language' semantics while preserving the text generation capability of the LLM}. We finetune the LLM with LoRA using a mixture of language and motion data in both unsupervised and supervised manners. Similar to LLaVA~\cite{liu2023llava}, we learn the complex relationship of language and motions with the paired text-motion datasets. Further, we ensure the text generation capability through the LLM's next-token prediction mechanism on text-only data. Compared with our conference MotionGPT~\cite{zhang2023motiongpt}, with the additional motion-language alignment stage, MotionGPT-2 offers improved semantic consistency and fine-grained control of human motions and textual descriptions.

\textbf{Stage 3: Fine-tuning LLM by Motion Instructions.}
% \label{sec: Finetuning LLM by Motion Instructions}
Instruction tuning enables LLMs to handle various motion generation and understanding tasks by asking LLM questions in different instructions. Hence, we devise a set of instructions that merge task descriptions with control signals and employ the efficient LoRA~\cite{hu2021lora} technique to fine-tune LLMs.

\begin{table*}[ht]
    \centering
    \small
    \setlength{\tabcolsep}{6.2pt}
    \caption{\textbf{Quantitative results of text-based motion generation on the HumanML3D dataset.} ``Real'' denotes the results computed with GT motions. ``$\rightarrow$'' indicates metrics that are better when closer to ``Real'' distribution. ``MultiModal Dist.'' denotes the Multi-Modality Distance. We conduct each evaluation 20 times, presenting the average metric and a 95\% confidence interval, with the top scores marked in bold.}
    \label{tab:humanml3d}
        \begin{tabular}{cccccccc}%
            \toprule
            \multirow{2}{*}{Model} & \multicolumn{3}{c}{R-Precision $\uparrow$} & \multirow{2}{*}{FID $\downarrow$} & \multirow{2}{*}{MultiModal Dist. $\downarrow$} & \multirow{2}{*}{Diversity $\rightarrow$} & \multirow{2}{*}{MultiModality $\uparrow$}\\
            & Top 1 &  Top 2 & Top 3 & \\
            \midrule
            Real                                            & 0.511$^{\pm.003}$          & 0.703$^{\pm.003}$          & 0.797$^{\pm.002}$          & 0.002$^{\pm.000}$          & 2.974$^{\pm.008}$          & 9.503$^{\pm.065}$          & —                           \\
            \midrule
            TM2T~\cite{guo2022tm2t}    & 0.424$^{\pm.003}$          & 0.618$^{\pm.003}$          & 0.729$^{\pm.002}$          & 1.501$^{\pm.046}$          & 3.467$^{\pm.008}$          & 8.589$^{\pm.058}$          & 2.424$^{\pm.093}$                           \\
            T2M~\cite{guo2022generating}                   & 0.457$^{\pm.002}$ & 0.639$^{\pm.003}$ & 0.740$^{\pm.003}$ & 1.067$^{\pm.002}$          & 3.340$^{\pm.008}$ & 9.188$^{\pm.002}$          & 2.090$^{\pm.083}$           \\
            MDM~\cite{tevet2023human}                      & 0.319$^{\pm.005}$          & 0.498$^{\pm.004}$          & 0.611$^{\pm.007}$          & 0.544$^{\pm.001}$          & 5.566$^{\pm.027}$          & 9.559$^{\pm.086}$ & \textbf{2.799}$^{\pm.072}$           \\
            MotionDiffuse~\cite{zhang2022motiondiffuse}                      & 0.491$^{\pm.001}$          & 0.681$^{\pm.001}$          & 0.782$^{\pm.001}$          & 0.630$^{\pm.001}$          & 3.113$^{\pm.001}$          & 9.410$^{\pm.049}$ &  1.553$^{\pm.042}$          \\
            MLD~\cite{chen2023executing}          & 0.481$^{\pm.003}$          & 0.673$^{\pm.003}$          & 0.772$^{\pm.002}$          & 0.473$^{\pm.013}$ & 3.169$^{\pm.010}$          & 9.724$^{\pm.082}$          & 2.413$^{\pm.079}$  \\
            T2M-GPT~\cite{zhang2023t2m}          & 0.491$^{\pm.003}$          & 0.680$^{\pm.003}$          & 0.775$^{\pm.002}$          & 0.116$^{\pm.004}$ & 3.118$^{\pm.011}$          & \textbf{9.761}$^{\pm.081}$          & 1.856$^{\pm.011}$  \\
            MoMask~\cite{guo2024momask}          & \textbf{0.521}$^{\pm.002}$          & \textbf{0.713}$^{\pm.002}$          & \textbf{0.807}$^{\pm.002}$          & \textbf{0.045}$^{\pm.002}$ & \textbf{2.958}$^{\pm.008}$          & 9.620$^{\pm.064}$          & 1.241$^{\pm.040}$  \\
            ReMoDiffuse~\cite{zhang2023remodiffuse}          & 0.510$^{\pm.005}$          & 0.698$^{\pm.006}$          & 0.795$^{\pm.004}$          & 0.103$^{\pm.004}$ & 2.974$^{\pm.016}$          & 9.018$^{\pm.075}$          & 1.795$^{\pm.043}$  \\
            AttT2M~\cite{zhong2023attt2m}          & 0.499$^{\pm.003}$          & 0.690$^{\pm.002}$          & 0.786$^{\pm.002}$          & 0.112$^{\pm.006}$ & 3.038$^{\pm.007}$          & 9.700$^{\pm.090}$          & 2.452$^{\pm.051}$  \\
            GraphMotion~\cite{jin2024act}          & 0.504$^{\pm.003}$          & 0.699$^{\pm.002}$          & 0.785$^{\pm.002}$          & 0.116$^{\pm.007}$ & 3.070$^{\pm.008}$          & 9.692$^{\pm.067}$          & 2.766$^{\pm.096}$  \\
            \midrule
            MotionGPT~\cite{zhang2023motiongpt}        & 0.364$^{\pm.005}$          & 0.533$^{\pm.003}$          & 0.629$^{\pm.004}$          & 0.805$^{\pm.002}$          & 3.914$^{\pm.013}$          & \textbf{9.972}$^{\pm.026}$ & \textbf{2.473}$^{\pm.041}$  \\
            MotionGPT~\cite{jiang2023motiongpt}                              & 0.492$^{\pm.003}$ & 0.681$^{\pm.003}$ & 0.733$^{\pm.006}$ & 0.232$^{\pm.008}$ & 3.096$^{\pm.008}$ & 9.528$^{\pm.071}$          & 2.008$^{\pm.084}$           \\
            MotionLLM~\cite{wu2024motionllm}        & 0.482$^{\pm.004}$          & 0.672$^{\pm.003}$          & 0.770$^{\pm.002}$          & 0.491$^{\pm.019}$          & 3.138$^{\pm.010}$          & 9.838$^{\pm.244}$ & —  \\
            MotionGPT-2~(Ours)        & \textbf{0.496}$^{\pm.002}$          & \textbf{0.691}$^{\pm.003}$          & \textbf{0.782}$^{\pm.004}$          & \textbf{0.191}$^{\pm.004}$          & \textbf{3.080}$^{\pm.013}$          & 9.860$^{\pm.026}$ & 2.137$^{\pm.022}$  \\
            \bottomrule
        \end{tabular}%
        \vspace{-0.3cm}
\end{table*}

% \section{Different From Current Related Methods}

% (1) Different from HumanTOMATO~\cite{lu2023humantomato}

% (2) Different from MotionGPT~\cite{jiang2023motiongpt}

% (3) Different from M3-GPT~\cite{luo2024m}

\section{Experiments}
\subsection{Experimental Setup}
% \subsection{Datasets and Evaluation Metrics}
% \textbf{Datasets}. We apply three publicly available datasets, HumanML3D~\cite{guo2022generating}, KIT-ML~\cite{plappert2016kit} and MotionX~\cite{lin2024motion}

The HumanML3D dataset~\cite{guo2022generating} stands as the largest available dataset focused solely on 3D body motion and associated textual descriptions. It comprises 14,616 motion clips paired with 44,970 meticulously annotated descriptions derived from a vocabulary of 5,371 unique words. These motion sequences are sourced from the AMASS~\cite{mahmood2019amass} and HumanAct12~\cite{guo2020action2motion} motion capture collections, showcasing a diverse array of human activities, including everyday tasks, sports, acrobatics, and artistic expressions. Each motion clip is linked with 3-4 descriptive texts. For training, one sentence is randomly chosen as the match, while for testing, the first text description is consistently used to evaluate model performance. The motion clips are down-sampled to 20 FPS and vary in duration from 2 to 10 seconds. The dataset is divided into training, validation, and test subsets, allocated in an 80\%, 5\%, and 15\% ratio, respectively, with no overlaps among them.

The KIT-ML~\cite{plappert2016kit} dataset is comprised of 3,911 motion sequences along with 6,278 textual descriptions. Each sequence is associated with one to four sentences, with the average description containing 9.5 words. This dataset merges selected elements from the KIT WholeBody Human Motion Database~\cite{mandery2015kit} and the CMU Graphics Lab Motion Capture Database~\cite{cmu}, with an emphasis on locomotion motions. The motion sequences in KIT-ML have been down-sampled to a frame rate of 12.5 fps.

\begin{table}[]
\centering
% \label{tab:motion-quality}
\small
\setlength{\tabcolsep}{0.8pt}
\caption{Assessment of motion generation across conditions. With initial/key tokens, MotionGPT($\dagger$) and MotionGPT-2($\ddagger$) show superior performance compared to the text-only version.}
\begin{tabular}{ccccccc}
\toprule
\multirow{2}{*}{Methods} & FID$\downarrow$        & Dist.$\downarrow$      & Diversity$\uparrow$     & FID$\downarrow$         & Dist.$\downarrow$      & Diversity$\uparrow$      \\
                         & \multicolumn{3}{c}{HumanML3D($\dagger$)} & \multicolumn{3}{c}{HumanML3D($\ddagger$)} \\ \midrule
Text-Only                & 0.567      & 3.775        & 9.006         & 0.191       & 3.080        & 9.860          \\
Text + Initial Pose      & 0.520      & 3.844        & 9.588         & 0.183       &  3.285            &    10.066            \\
Text + Last Pose         & 0.591      & 3.718        & 9.251         & 0.358       &  3.673        &  9.582              \\
Text + Random Pose       & 0.367      & 3.598        & 9.176         & 0.182          &  3.031            & 10.102               \\ \midrule
                         & \multicolumn{3}{c}{KIT-ML($\dagger$)}    & \multicolumn{3}{c}{KIT-ML($\ddagger$)}    \\ \midrule
Text-Only                & 0.597      & 3.394        & 10.540         & 0.614       & 3.164        & 11.256         \\ 
Text + Initial Pose      & 0.664      & 3.445        & 10.390         & 0.756       & 3.362        & 11.053         \\
Text + Last Pose         & 0.856      & 3.336        & 10.580         & 0.784       & 3.483        & 11.460         \\
Text + Random Pose       & 0.671      & 3.411        & 10.760         & 0.807       & 3.173        & 11.447               \\ \bottomrule
\end{tabular}
\vspace{-0.2cm}
\label{tab:motion-quality}
\end{table}

The Motion-X~\cite{lin2024motion} dataset is currently the largest whole-body expressive motion repository, comprising 95,642 high-fidelity 3D motion sequences based on the SMPL-X model~\cite{pavlakos2019expressive}, paired with corresponding pose descriptions and semantic labels for each sequence. The Motion-X dataset gathers 15K monocular videos from various online sources and public video dataset, capturing a wide range of scenarios such as daily actions, sports activities, and many domain-specific scenes, with 13.7M frame-level 3D whole-body pose annotations. In this paper, we standardize the Motion-X dataset by selecting 52 joints from the human body and hands.\footnote{More experiments results and visualizations on the KIT-ML benchmark are available in the \textit{Supplementary Materials}.}
\vspace{-0.3cm}

\begin{table*}[]
\centering
\small
\setlength{\tabcolsep}{5.8pt}
\caption{Experiments of motion captioning task on the HumanML3D~\cite{guo2022generating} benchmark. Results marked with * are from MotionGPT~\cite{jiang2023motiongpt}, and were computed using unprocessed ground truth texts for linguistic metrics.}
\begin{tabular}{ccccccccccc}
\toprule
\multirow{2}{*}{Methods} & \multicolumn{3}{c}{R Precision$\uparrow$} & \multirow{2}{*}{MultiModal Dist.$\downarrow$} & \multirow{2}{*}{Length$_\text{avg}$$\uparrow$} & \multirow{2}{*}{Bleu\@1$\uparrow$} & \multirow{2}{*}{Bleu\@4$\uparrow$} & \multirow{2}{*}{Rouge$\uparrow$} & \multirow{2}{*}{Cider$\uparrow$} & \multirow{2}{*}{BertScore$\uparrow$} \\
                         & Top1      & Top2     & Top3     &                          &                            &                         &                         &                        &                        &                            \\ \midrule
Real Desc                & 0.523     & 0.725    & 0.828    & 2.901                    & —                          & —                       & —                       & —                      & —                      & —                          \\
RAEs                     & 0.100     & 0.188    & 0.261    & 6.337                    & —                          & 33.3                    & 10.2                    & 37.5                   & 22.1                   & 10.7                       \\
Seq2Seq(Att)             & 0.436     & 0.611    & 0.706    & 3.447                    & —                          & 51.8                    & 17.9                    & 46.4                   & 58.4                   & 29.1                       \\
SeqGAN                   & 0.332     & 0.457    & 0.532    & 4.895                    & —                          & 47.8                    & 13.5                    & 39.2                   & 50.2                   & 23.4                       \\ \midrule
TM2T*                    & 0.516     & 0.720    & 0.823    & 2.935                    & 10.67                      & \textbf{48.9}                    & 7.00                    & \textbf{38.1}                   & 16.8                   & 32.2                       \\
MotionGPT*                & 0.543     & —        & 0.827    & 2.821                    & 13.04                      & 48.2                    & 12.5                    & 37.4                   & 29.2                   & 32.4                       \\
MotionGPT-2 (Ours)        & \textbf{0.558}     &  \textbf{0.738}  &  \textbf{0.838}  &  \textbf{2.767}                      &  \textbf{15.27}                   &    48.7                     &  \textbf{13.8}                       &  37.6                      &  \textbf{29.8}                    &   \textbf{32.6}                         \\ \bottomrule
\end{tabular}
\label{tab:motion captioning}
\vspace{-0.1cm}
\end{table*}

\begin{figure}[ht]
    \centering
    \includegraphics[width=\linewidth]{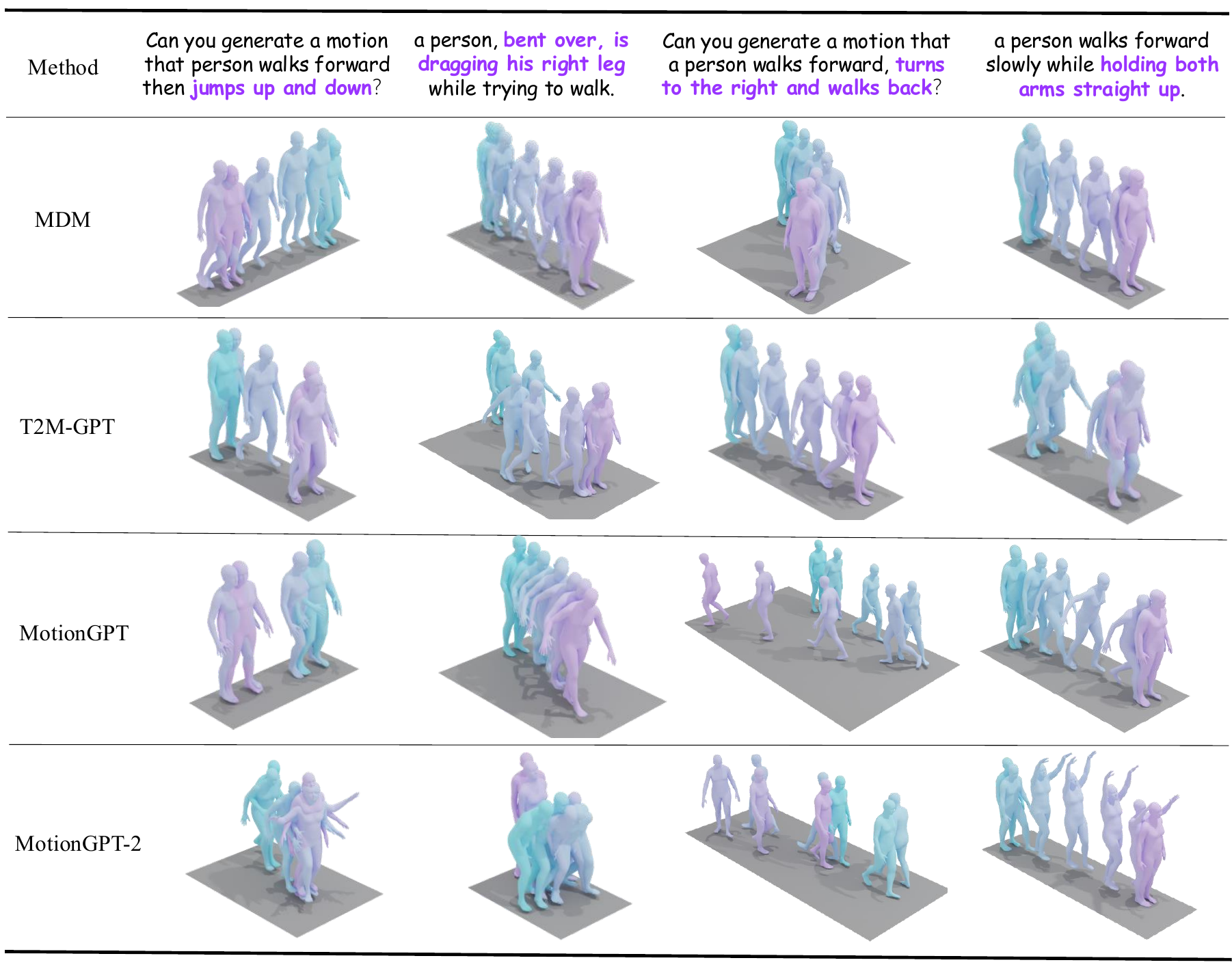}
    \caption{
    Showcase of visualization results for the text-based motion generation task using the HumanML3D~\cite{guo2022generating} dataset. We compare our MotionGPT-2 with the state-of-the-art method, \textit{i.e.}, MDM~\cite{tevet2023human}, T2M-GPT~\cite{zhang2023t2m}, MotionGPT~\cite{zhang2023motiongpt}. 
    }
    \label{fig:Compare_with_sota}
    \vspace{-0.55cm}
\end{figure}

% \vspace{-0.3cm}
\subsection{Implementation Details}
\textbf{Motion Data Pre-processing.}
We follow the dataset pre-processing procedures outlined in \cite{guo2022generating,zhang2023t2m,zhang2023motiongpt}. The raw 3D motion coordinates are first aligned with a default human skeletal model. The Y-axis is then set perpendicular to the ground, allowing individuals to face the Z+ direction. These coordinates are then processed into motion features, including foot contact, global rotations and translations, local joint positions, velocities, and 6D rotations. The final dimensions are 263 for the HumanML3D dataset, 251 for the KIT-ML dataset, and 623 for the Motion-X dataset. For SMPL-based motion representations, the maximum motion length is set to 196, with minimums of 40 for HumanML3D and 24 for KIT-ML. In SMPL-X-based Motion-X dataset, the maximum motion length frames is 300, with a minimum of 40.

\begin{table*}[t]
    \centering
    \small
    \setlength{\tabcolsep}{7pt}
    \caption{Ablations on the effects of LLM types and scales on text-based motion generation, evaluated on the HumanML3D~\cite{guo2022generating} benchmark. In addition to full fine-tuning of the T5-base model, LoRA is used for optimizing other decoder-only LLMs. }
    \label{tab:humanml3d_llm}
        \begin{tabular}{ccccccccc}%
            \toprule
            \multirow{2}{*}{Model} & \multirow{1}{*}{Trainable} & \multicolumn{3}{c}{R-Precision $\uparrow$} & \multirow{2}{*}{FID $\downarrow$} & \multirow{1}{*}{MultiModal} & \multirow{2}{*}{Diversity $\rightarrow$} & \multirow{2}{*}{MultiModality $\uparrow$}\\
            & Parameters & Top 1 &  Top 2 & Top 3 &  & Dist. $\downarrow$ \\
            \midrule
            Real                                 &     —      & 0.511          & 0.703          & 0.797          & 0.002          & 2.974          & 9.503          & —                           \\ \midrule
            T5-base (Pretrain)     &  Full-Finetune  & 0.314          & 0.457          & 0.554          & 0.493          & 4.662          & 9.634 & 1.793  \\
            \rowcolor{gray!20}
            T5-base (Finetune)     &  Full-Finetune   & 0.417 & 0.581 & 0.668 & 0.148 & 3.989 & 9.961  &  1.985          \\
            Gemma-2b-It (Pretrain)   &   34M   & 0.385          & 0.546          & 0.643          & 0.207          & 4.559       & 9.733 & 2.351  \\
            \rowcolor{gray!20}
            Gemma-2b-It (Finetune)   &   34M    &  0.436        &  0.600        &    0.697       &    0.228      &  3.589      & 10.081 & 2.269  \\
            Gemma-7b-It (Pretrain)   &  101M    & 0.404          & 0.575          & 0.673         & 0.219          & 3.768          & 9.964 & 2.364  \\
            \rowcolor{gray!20}
            Gemma-7b-It (Finetune)   &  101M     & 0.446          & 0.622          & 0.715          & 0.177          & 3.545          & 9.652 &  2.198 \\
            LLaMA3-8B (Pretrain)    &  89M   & 0.438          & 0.619         & 0.717          & 0.314          & 3.514          & 10.010 & 2.252  \\
            \rowcolor{gray!20}
            LLaMA3-8B (Finetune)    &   89M  & 0.482 & 0.668 & 0.760         & 0.282          & 3.288          & 10.212 & 2.261 \\
            LLaMA3.1-8B (Pretrain)  &  89M      & 0.456          & 0.630          & 0.732          & 0.291          & 3.417          & 9.884 & 2.145  \\
            \rowcolor{gray!20}
            LLaMA3.1-8B (Finetune)   &   89M    & \textbf{0.496}          & \textbf{0.691}          & \textbf{0.782}          & \textbf{0.191}         & \textbf{3.080}          & 9.860 & 2.137  \\
            
            \bottomrule
        \end{tabular}
        \vspace{-0.3cm}
\end{table*}

\textbf{Training Details.}
In the following experiments, we adopt the decoder-only LLaMA 3.1-8B~\cite{touvron2023llama} model as the foundational LLM, keeping its parameters frozen while applying fine-tuning through the LoRA~\cite{hu2021lora} method. The motion tokenizer is trained over 1,200 epochs, with an initial learning rate of $1\times10^{-4}$. We set the mini-batch size to 256 and use the AdamW optimizer~\cite{loshchilov2017decoupled}, with a weight decay of $1\times10^{-5}$ for model optimization. Following previous studies~\cite{zhang2023t2m,jiang2023motiongpt}, we set the codebook $\mathcal{B}_m\in \mathbb{R}^{512\times 512}$ for motion VQ-VAE and the codebook $\mathcal{B}_b\in \mathbb{R}^{512\times 512}$ and $\mathcal{B}_h\in \mathbb{R}^{512\times 512}$ for Part-Aware VQ-VAE. The motion encoder applies a temporal down-sampling rate of 4. Our MotionGPT-2 leverages a learning rate of 2$\times$10$^{-4}$ during the pre-training phase and 1$\times$10$^{-4}$ in the instruction tuning phase of the unified motion-language learning process. The mini-batch size is 32. The pre-trained LLM undergoes 100 epochs in the \textit{Motion-Language Alignment} phase and 50 epochs during the \textit{Instruction Tuning} phase. MotionGPT-2 and the re-implemented models are built using PyTorch, with all experiments on 4 NVIDIA 80G A100 GPUs.

\vspace{-0.45cm}
\subsection{Results on Human Motion Generation} 
\textbf{Text-based Body-only Motion Generation}. The quantitative results of motion quality are depicted in Table~\ref{tab:humanml3d}. Among these, Table~\ref{tab:humanml3d} provide quantitative comparisons on the SMPL-based HumanML3D~\cite{guo2022generating} dataset and the KIT-ML~\cite{mandery2015kit} dataset. The results presented reflect the performance of MotionGPT-2, which has been pre-trained across multiple motion-related tasks and subsequently fine-tuned for the specific text-based motion generation task. By tuning mere 1\% of LLM parameters, our general-purpose MotionGPT-2 exhibits a performance that is competitive with state-of-the-art approaches. Compared to our previous version, the MotionGPT-2 yields a significant \textbf{10.4}\% improvement in R-Precision Top-3 and a \textbf{0.254} reduction in FID score on HumanML3D. Through the tailored \textit{Motion-Language Alignment} stage, our MotionGPT-2 exhibits greater semantic consistency with the textual description, improving interpretation of body language semantics. Further, compared to other LLM-based methods, our MotionGPT-2 fine-tunes the LLM and relate world knowledge to 3D human motions, achieving superior generation beyond existing solutions.

\begin{figure}[t]
    \centering
    \includegraphics[width=0.5\textwidth]{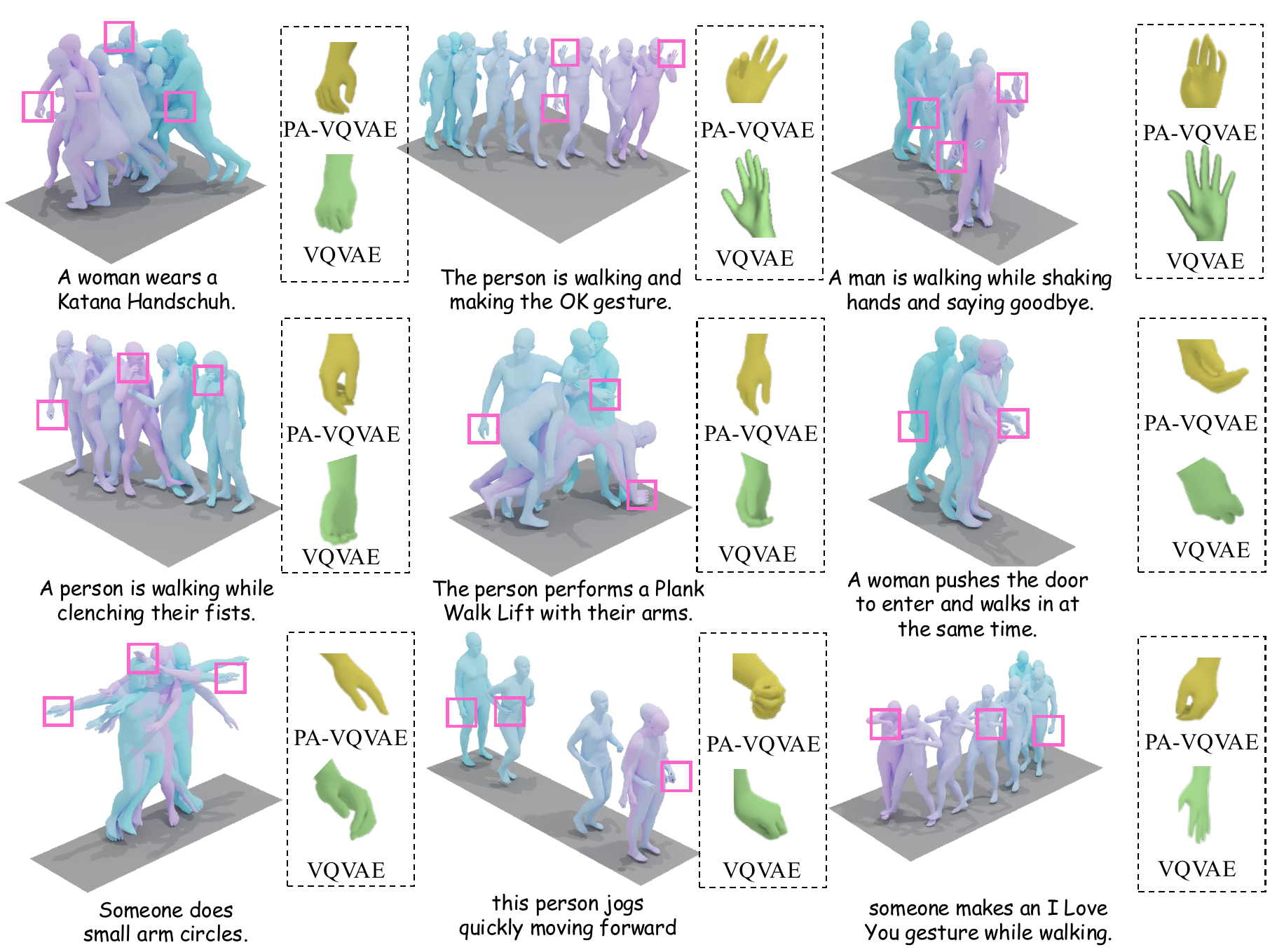}
    \caption{
    Qualitative results of our method on Motion-X~\cite{lin2024motion}. Utilizing world knowledge of LLMs, our PA-VQVAE demonstrates the capability to effectively capture lifelike hand interactions.
    % \textit{e.g.}, \textit{making the OK gesture}, \textit{plays the piano}, \textit{saying goodbye}.
    } % , \textit{e.g.}, \textit{making the OK gesture}, \textit{plays the piano}, \textit{saying goodbye}
    \label{fig:motionx}
    \vspace{-0.2cm}
\end{figure}

\begin{table}[]
\centering
\small
\setlength{\tabcolsep}{8.4pt}
\begin{tabular}{ccccc}
\toprule
\multirow{2}{*}{Initial token} & Recon         & Vel           & Recon           & Vel           \\
                               & \multicolumn{2}{c}{MotionGPT} & \multicolumn{2}{c}{MotionGPT-2} \\ \midrule %\cline{2-5} 
Text-only                      & 24.70         & 1.095         &  19.19          & 0.846              \\
Text + Initial poses           & \textbf{13.78}         & \textbf{0.549}         &  \textbf{7.59}               &     \textbf{0.328}          \\ \midrule
Last token                     & \multicolumn{2}{c}{MotionGPT} & \multicolumn{2}{c}{MotionGPT-2}   \\ \midrule %\cline{2-5} 
Text-only                      & 19.70         & 1.172         &  11.77               &  0.735             \\
Text + Last poses              & \textbf{6.831}         & \textbf{0.397}         &   \textbf{4.376}              &   \textbf{0.291}            \\ \midrule
Key tokens                     & \multicolumn{2}{c}{MotionGPT} & \multicolumn{2}{c}{MotionGPT-2}   \\ \midrule %\cline{2-5} 
Text-only                      & 8.035         & 3.813         &  6.591          &  1.932             \\
Text + Random poses            & \textbf{5.383}         & \textbf{2.423}         &  \textbf{4.139}               &   \textbf{1.055}            \\ \bottomrule
\end{tabular}
\caption{Evaluation of the consistency with pose control conditions on the HumanML3D~\cite{guo2022generating} test sub-set using the pre-trained LLaMA 3.1-8B model.} % 
\label{tab:pose-consitency}
\vspace{-0.4cm}
\end{table}

As shown in Fig.~\ref{fig:Compare_with_sota}, we conduct visualization experiments on the text-based motion generation task to vividly illustrate the capabilities of our MotionGPT-2 model. In Fig.~\ref{fig:Compare_with_sota}, we observe that the diffusion-based MDM~\cite{tevet2023human} method generates fewer semantic motions aligned with the provided descriptions. By utilizing discrete motion tokens, T2M-GPT~\cite{zhang2023t2m} can better learn motion patterns and semantics which leads to more coherent and contextually rich motions. Our conference version MotionGPT~\cite{zhang2023motiongpt}, with fine-tuned LLMs, generalizes well to more diverse and complex text prompts. In contrast to MotionGPT, our MotionGPT-2 extends motion vocabulary to original LLMs for jointly modeling motion and language representations in a unified space. 
% As demonstrated by the additional visualization results in Fig.~\ref{fig:Addition_vis_humanml3d}, MotionGPT-2 shows high-fidelity motion and strong text prompt matching, which validates the effectiveness of our proposed method.

\textbf{Multiple Signal Controlled Body-only Motion Generation}. Besides text inputs, MotionGPT-2 is capable of incorporating human poses as an additional control modality, with the resulting motion quality detailed in Table~\ref{tab:motion-quality}. Introducing extra controls, such as initial, last, or key poses, does not degrade motion quality. In fact, MotionGPT-2 shows superior performance when initial or key tokens are provided, achieving FID scores of \textbf{0.183} or \textbf{0.182}, an improvement over the text-only model's 0.191 on the HumanML3D benchmark, demonstrating its flexibility in handling diverse control modalities.In fact, MotionGPT-2 shows superior performance when initial or key tokens are provided, achieving FID scores of \textbf{0.183} or \textbf{0.182}, an improvement over the text-only model's 0.191 on the HumanML3D benchmark. In spite of this, MotionGPT-2's performance is still notable, reinforcing its proficiency in generating high-quality and diverse motions across a range of control conditions. As demonstrated in Fig.~\ref{fig:multiple-control}, the motions produced by our model align closely with the specified poses and consistently follow the textual descriptions.

\begin{table*}[]
    \centering
    \small
    \caption{Comparisons between separate training for each task and joint training for multiple tasks on HumanML3D~\cite{guo2022generating} test set using the LLaMA 3-8B model. We use orange and green markings to represent decrements and improvements in the metric, respectively. Joint training can achieve better performance for all tasks.} %  
    \begin{tabular}{l|c|cccccc}
         \toprule
             \multirow{2}{*}{Task} & Training & \multirow{2}{*}{FID $\downarrow$} & \multirow{2}{*}{MultiModal Dist.$\downarrow$} & \multicolumn{3}{c}{R-Precision $\uparrow$} & \multirow{2}{*}{Diversity $\uparrow$} \\
            & Strategy & & & Top-1 & Top-2 & Top-3 & \\
        \midrule
            Text & \multirow{4}{*}{Separate} &  0.523 & 3.627 & 0.358  & 0.514 & 0.604 & 9.108  \\
            \quad + Initial token &   &  0.483 & 3.489 & 0.378 &  0.549   & 0.647  & 9.614  \\
            \quad + Last token &  & 0.974 & 4.208  &  0.339 & 0.501  & 0.598  & 9.598    \\
            \quad + Key tokens &  & 0.428 & 3.276  & 0.424  &  0.617 & 0.697  & 9.929  \\
        \bottomrule
        \toprule
            Text & \multirow{4}{*}{Joint} & 0.482(\textcolor[RGB]{255,153,51}{-0.041}) & 3.295(\textcolor[RGB]{255,153,51}{-0.332}) & 0.419(\textcolor[RGB]{87,219,150}{+0.061}) & 0.597(\textcolor[RGB]{87,219,150}{+0.083}) & 0.683(\textcolor[RGB]{87,219,150}{+0.079}) & 9.422(\textcolor[RGB]{87,219,150}{+0.314}) \\
            \quad + Initial token & & 0.454(\textcolor[RGB]{255,153,51}{-0.029}) & 3.173(\textcolor[RGB]{255,153,51}{-0.316}) & 0.434(\textcolor[RGB]{87,219,150}{+0.056}) & 0.613(\textcolor[RGB]{87,219,150}{+0.064}) & 0.710(\textcolor[RGB]{87,219,150}{+0.063}) & 9.573(\textcolor[RGB]{255,153,51}{-0.041}) \\
            \quad + Last token & & 0.507(\textcolor[RGB]{255,153,51}{-0.467}) & 3.860(\textcolor[RGB]{255,153,51}{-0.348}) & 0.427(\textcolor[RGB]{87,219,150}{+0.088}) & 0.608(\textcolor[RGB]{87,219,150}{+0.107}) & 0.723(\textcolor[RGB]{87,219,150}{+0.125}) & 9.688(\textcolor[RGB]{87,219,150}{+0.090}) \\
            \quad + Key tokens &  & 0.406(\textcolor[RGB]{255,153,51}{-0.022}) & 3.459(\textcolor[RGB]{255,153,51}{-0.183}) & 0.445(\textcolor[RGB]{87,219,150}{+0.021}) & 0.616(\textcolor[RGB]{255,153,51}{-0.001}) & 0.722(\textcolor[RGB]{87,219,150}{+0.025}) & 9.987(\textcolor[RGB]{87,219,150}{+0.058}) \\
        \bottomrule
    \end{tabular}
    \label{tab:separate}
    \vspace{-0.25cm}
\end{table*}

\begin{figure}[]
    \centering
    \includegraphics[width=\linewidth]{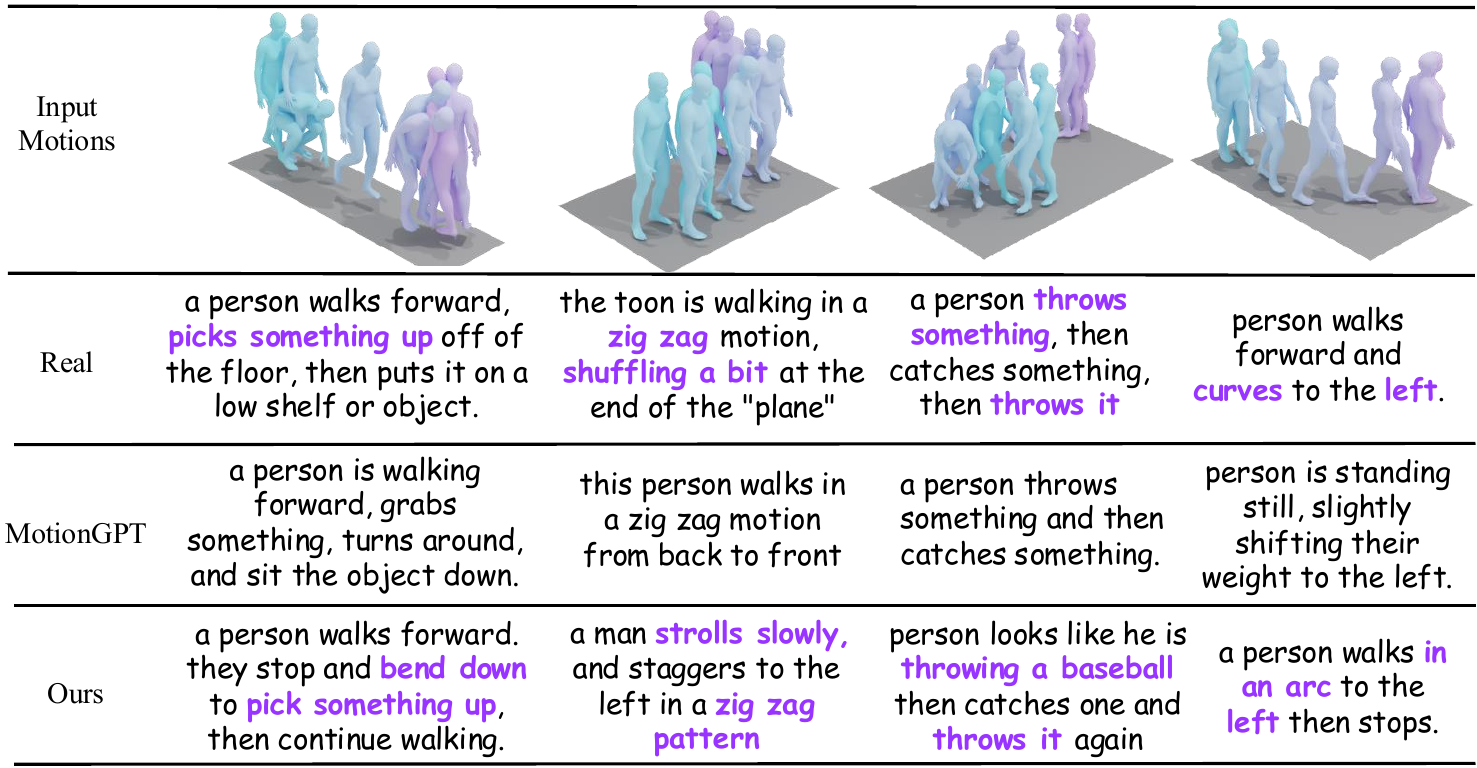}
    \caption{
    Comparison of other methods on motion captioning task. The results demonstrate that our MotionGPT-2 generates more conceptually and semantically rich motion descriptions. Specific words are marked to highlight the semantic similarity of the generated captions and the real one. 
    }
    \label{fig:motion_captioning}
    \vspace{-0.3cm}
\end{figure}

\begin{table}[]
\centering
\small
\setlength{\tabcolsep}{1pt}
\caption{Evaluation of motion prediction and in-betweening on part of the AMASS~\cite{mahmood2019amass}, considering only motion data. FID reflects the quality of the generated motions, while Diversity quantifies the motion variability within each condition. }
% ADE and FDE represent the distance between generated joint positions and the ground truth.} % 
\begin{tabular}{@{}cccccccccc@{}}
\toprule
\multirow{2}{*}{Methods} & 
\multicolumn{4}{c}{Motion Prediction }& &
\multicolumn{3}{c}{Motion In-between  }\\
\cmidrule(lr){2-5} \cmidrule(lr){7-9}
& $\text{FID}\downarrow $  
& Diversity$\uparrow$            
& ADE$\downarrow$
& FDE$\downarrow$
&& $\text{FID}\downarrow $  
& Diversity$\uparrow$            
& ADE$\downarrow$
\\ 
\toprule
Real & 0.002 & 9.503 & — & — && 0.002 & 9.503 & —
\\\midrule
% T2M-GPT\cite{}& \textcolor[RGB]{255,153,51}{TODO} &\textcolor[RGB]{255,153,51}{TODO}&\textcolor[RGB]{255,153,51}{TODO}&\textcolor[RGB]{255,153,51}{TODO} &-&-&-\\
MDM & 6.031 & 7.813 & 5.446 & 8.561 && 2.698 & 8.420& 3.787
\\
% \midrule
MotionGPT
& 0.905 & 8.972 & 4.745& 6.040 && \textbf{0.214} & \textbf{9.560} & 3.762 \\
MotionGPT-2
& $\textbf{0.537}$ &$\textbf{9.414}$ & $\textbf{4.512}$& $\textbf{5.823}$ &&0.408&9.327&$\textbf{3.704}$
\\ \bottomrule
\end{tabular}%
\label{tab:Motion Completion}
\vspace{-0.32cm}
\end{table}

\begin{figure*}[] 
\begin{center} 
\centerline{\includegraphics[width=1\linewidth]{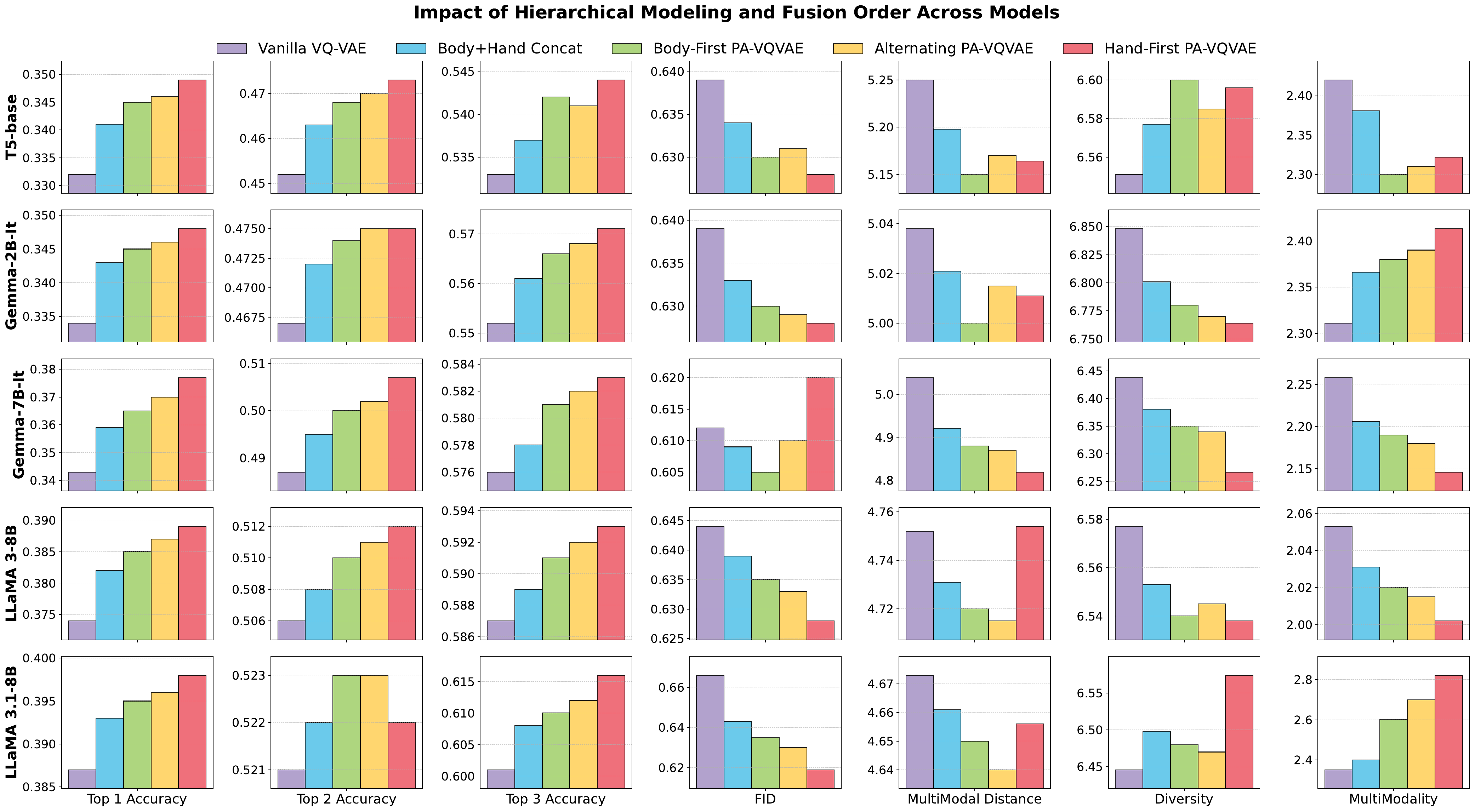}} 
\caption{\textbf{Impact of Hierarchical Modeling and Fusion Order on Motion Generation Performance.} This ablation study across five LLM backbones demonstrates that our Hand-First PA-VQVAE consistently outperforms Vanilla VQ-VAE, Body+Hand Concat, Body-First, and Alternating variants across key metrics (Top-1/2/3 Accuracy, FID, Diversity, MultiModality).} 
\label{fig:pa-vqvae-ablation-all-models} 
\end{center} 
\vspace{-0.8cm}
\end{figure*}

\textbf{Holistic Motion Generation}. In this part, we focus on the holistic motion generation task. As shown in Figure~\ref{fig:pa-vqvae-ablation-all-models}, we evaluate various LLMs on the SMPL-X-based Motion-X~\cite{lin2024motion} dataset. Our tailored PA-VQVAE consistently outperforms the original VQVAE across multiple scales and types of LLMs, demonstrating the effectiveness of our innovative motion discretization framework. For instance, compared to its VQVAE counterpart, Part-Aware VQVAE (LLaMA 3.1-8B) achieves an improvement in Top-1 R-Precision from 0.332 to 0.349, and a decrease in FID from 0.666 to 0.619. The superior performance of Part-Aware VQVAE over the original VQVAE can be attributed to its more fine-grained discretization representation and hierarchical modeling capability for human motion. Moreover, utilizing distinct motion vocabularies for body and hand reduces the ambiguity, where similar actions could be represented by the same token. As shown in Fig.~\ref{fig:motionx}, our MotionGPT-2 is capable of generating vivid motions that accurately correspond to the given text descriptions.

\textbf{Perceptual Studies}. We assess visual realism and semantic fidelity of generated 3D motions by rendering 2,000 representative clips from multiple camera views and evaluating them with Qwen2.5-VL-72B~\cite{bai2025qwen2}, which scores each sample across five dimensions: \textbf{Naturalness}, \textbf{Semantic Consistency}, \textbf{Fluency}, \textbf{Detail}, and \textbf{Overall Preference} (definitions in the supplementary material). In Figure~\ref{fig:perceptual_comparison}, MotionGPT-2 consistently surpasses three strong baselines (MotionGPT, Gemma-2B-It, Gemma-7B-It) across all dimensions, demonstrating its clear perceptual superiority. MotionGPT-2 attains the highest mean scores in Naturalness (3.84) and Fluency (3.45), reflecting physically coherent and temporally smooth motion generation enabled by Part-Aware VQ-VAE. Its Semantic Consistency (3.74) highlights accurate grounding between texts and motion dynamics through instruction tuning. In Fine-Grained Detail (3.16), the model captures subtle hand and body gestures, a long-standing challenge for prior methods. Finally, the Overall Preference (4.11)—obtained via pairwise comparison—confirms MotionGPT-2's clear human-perceived advantage, with over 60\% of evaluators preferring its outputs. Together, these results provide strong human-centered evidence that the proposed PA-VQVAE and LLM designs of MotionGPT-2 enhance motion quality, realism, and semantic fidelity.

\textbf{User Study for Fine-Grained Hand Motion Evaluation}. We conduct a small-scale user study involving 10 participants with backgrounds in computer vision or human motion analysis. Each participant is presented with 20 paired video clips generated by the vanilla VQ-VAE and our Part-Aware VQ-VAE under identical text prompts (e.g., “making the OK gesture”, “playing the piano”). In each pair, the participants are asked to select the sequence that appears more natural in terms of fine-grained hand articulations. Across the total of 200 pairwise comparisons, 81.5\% of the selections favor the outputs of PA-VQVAE. This strong and consistent preference indicates that our hierarchical design not only improves objective reconstruction metrics but also yields perceptible gains in the naturalness and plausibility of hand motions. These results provide compelling subjective evidence, complementing our quantitative evaluations, and directly address the reviewer’s suggestion to assess hand motion quality through user studies.

\begin{figure}[] 
\begin{center} 
\centerline{\includegraphics[width=\linewidth]{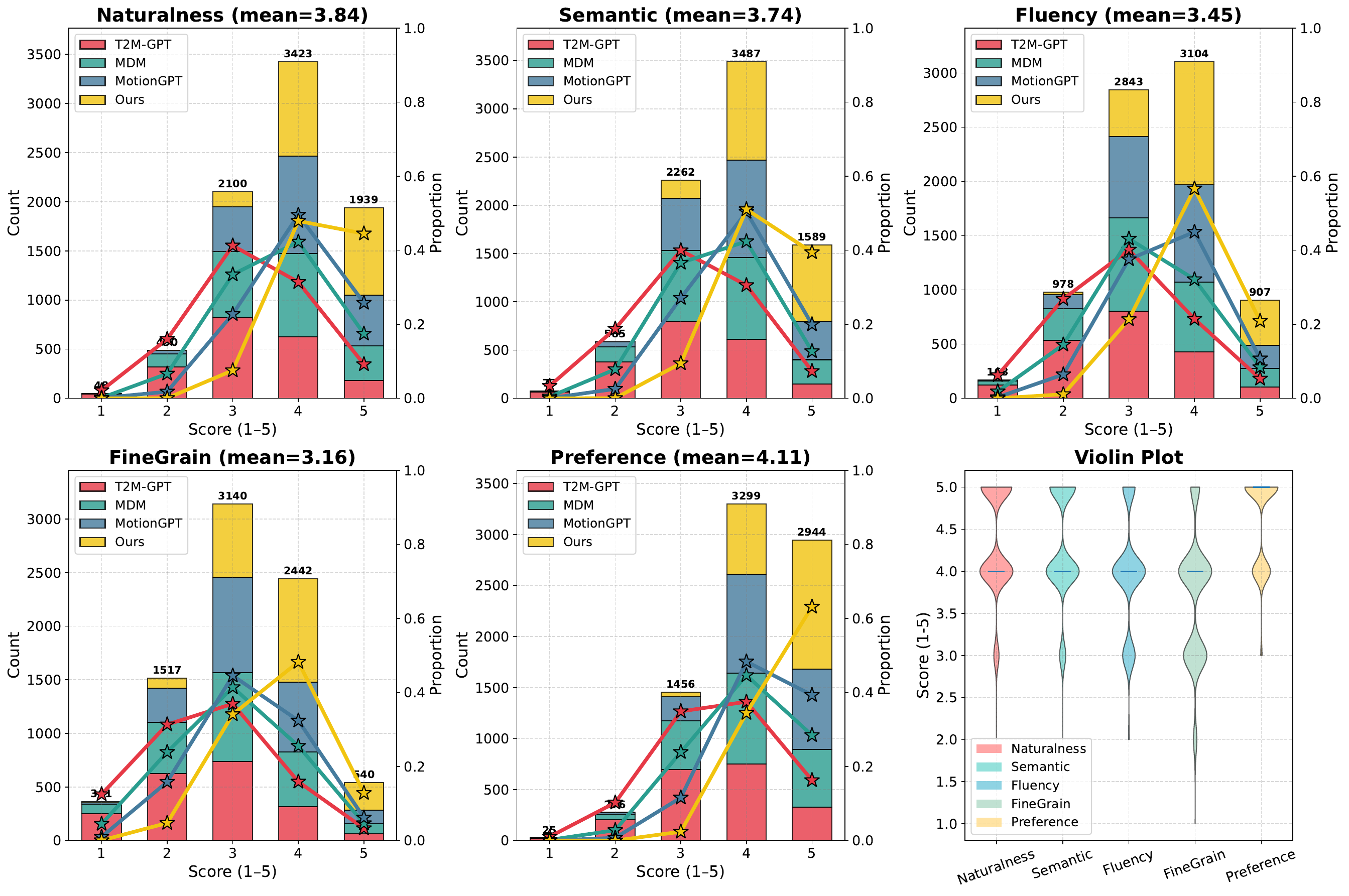}} 
\caption{\textbf{Subjective Assessment of Motion Quality}. A subjective assessment of our model’s performance compared to T2M-GPT, MDM, and MotionGPT across five qualitative metrics.} 
\label{fig:perceptual_comparison} 
\end{center} 
\vspace{-0.8cm}
\end{figure}

\begin{figure}[ht] 
\begin{center} 
\centerline{\includegraphics[width=1\linewidth]{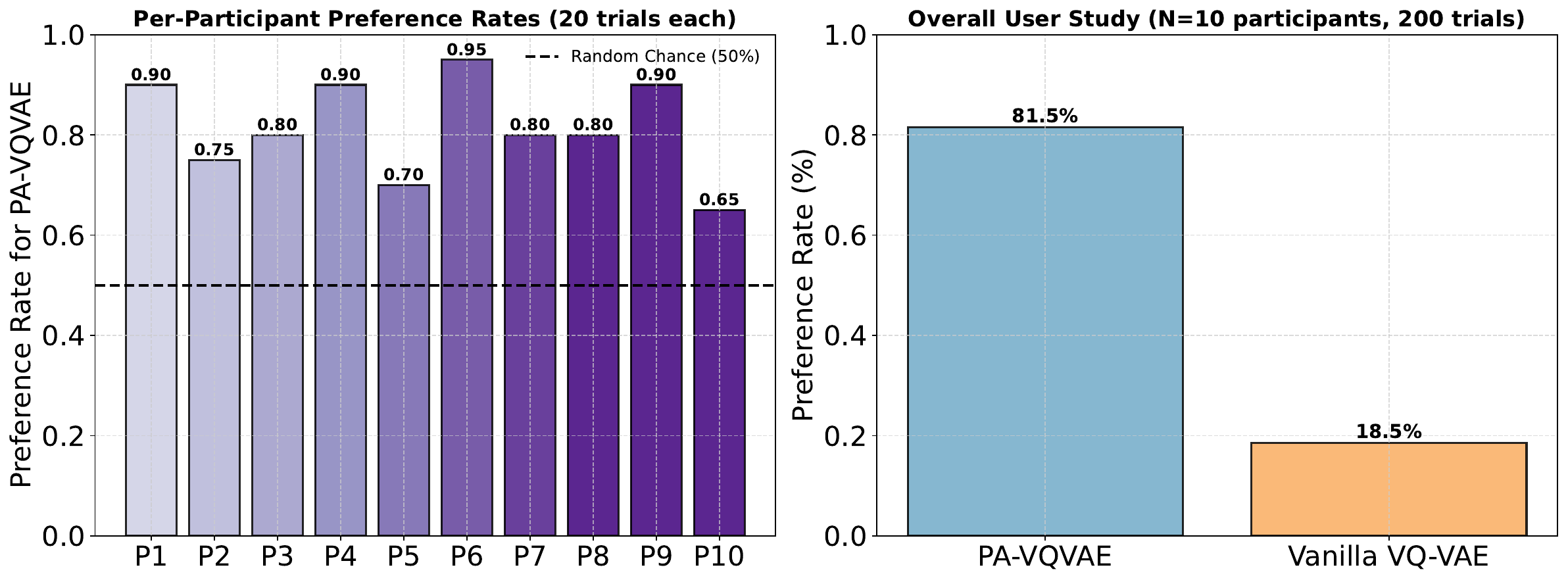}} 
\caption{\textbf{Subjective Evaluation of Hand Motion Fidelity}. (\textbf{Left}) Individual preference rates from a 10-participant study comparing PA-VQVAE and Vanilla VQ-VAE under identical prompts. (\textbf{Right}) Aggregated outcomes reveal an 81.5\% overall preference for PA-VQVAE across 200 trials.} 
\label{fig:per_participant_preference_rates_purple} 
\end{center} 
\vspace{-0.8cm}
\end{figure}

\vspace{-0.3cm}
\subsection{Results on Motion Captioning}
Compared to the text-to-motion task, the motion-to-text task involves generating a textual description from a given human motion sequence. As in \cite{jiang2023motiongpt}, we rely on ground truth descriptions for more accurate assessment.  Table~\ref{tab:motion captioning} presents the results of quantitative evaluation for motion-to-text translation on the HumanML3D dataset. The comparisons in Table~\ref{tab:motion captioning} shows that our proposed MotionGPT-2 outperforms recent works in generating text descriptions for the given motions. The generated language descriptions deliver substantial improvements in both linguistic quality (BLEU~\cite{papineni2002bleu} and BertScore~\cite{zhang2019bertscore}) and the precision of motion retrieval (R-Precision). By fine-tuning LLMs, our MotionGPT-2 emerges as a specialized tool endowed with extensive world knowledge, thereby enhancing its capacity to interpret human motion.

In Fig.~\ref{fig:motion_captioning}, we provide further examples of translating motion to text using the HumanML3D dataset. All methods are evaluated under the same training and inference conditions on HumanML3D. Compared to MotionGPT~\cite{jiang2023motiongpt}, our MotionGPT-2, which leverages LLM-interpretable motion tokens, is capable of generating more conceptual and semantically rich motion descriptions. For instance, it can produce descriptions such as ``\textit{walk in an arc shape}'' and ``\textit{throwing a baseball}'', offering more intuitive understanding of the given motion motions.

% \begin{figure*}[]
%     \centering
%     \includegraphics[width=1\textwidth]{Addition_vis_humanml3d.pdf}
%     \caption{
%     More text-based human motion samples generated by our proposed MotionGPT-2 (with LLaMA 3.1-8B) using texts from the HumanML3D test set. Our method effectively generates a diverse range of dynamic and imaginative motions,  \textit{e.g.}, ``\textit{acting like a squirrel}'', ``\textit{walking in an S-shaped pattern}'', and ``\textit{dancing around}''.
%     }
%     \label{fig:Addition_vis_humanml3d}
%     \vspace{-0.25cm}
% \end{figure*}

\vspace{-0.35cm}
\subsection{Results on Generalized Motion Completion}
Following~\cite{jiang2023motiongpt}, we classify both motion prediction and in-betweening together as generalized motion completion in Table~\ref{tab:Motion Completion}. For the motion prediction task~\cite{ma2022multi,yuan2020dlow,zhang2021we}, only the first 20\% of the sequence is used as the conditioning input, while approximately 50\% of the motion is intentionally masked at random to assist in the completion process. Similar to text-based motion generation, as shown in Table~\ref{tab:Motion Completion}, we also fine-tune the proposed MotionGPT-2 specifically for this specific task and utilize FID, ADE, and FDE as metrics. The unified motion-language framework of the MotionGPT-2 leverages contextual information of fine-tuned LLM to understand in-depth motion dynamics. Compared to \cite{jiang2023motiongpt,tevet2023human}, our MotionGPT-2 demonstrates a remarkable capability to generate contextually appropriate motion completions.

% \begin{table}[]
% \setlength{\tabcolsep}{3.1pt}
%     \centering
%     \small
%     \caption{Evaluation of text-based human motion generation using the LLaMA 3-8B model with various prompts on the HumanML3D~\cite{guo2022generating} test subset.}

%     \begin{tabular}{c|cccccc}
%          \toprule
%              \multirow{2}{*}{Prompts} & \multirow{2}{*}{FID $\downarrow$} & \multirow{1}{*}{MultiModal} & \multicolumn{3}{c}{R-Precision $\uparrow$} & \multirow{2}{*}{Diversity $\uparrow$} \\
%              %\cline{2-4}
%             & &  Dist. $\downarrow$ & Top-1 & Top-2 & Top-3 & \\
%         \midrule
%             $V_1$ & 4.196 & 5.275 & 0.357 & 0.542 & 0.658 & 8.110 \\
%             $V_2$ & 2.692 & 4.573 & 0.418 & 0.603 & 0.719 & 8.534 \\
%             $V_0$ (Ours) & {\bf 0.191} & {\bf 3.080} & {\bf 0.482} & {\bf 0.669} & {\bf 0.760} & {\bf 9.860} \\
%         \bottomrule
%     \end{tabular}
%     \label{tab:prompts}
%     \vspace{-0.45cm}
% \end{table}

\vspace{-0.4cm}
\subsection{Ablation Study}

\textbf{Capability of Pre-trained LLM.} As shown in Table~\ref{tab:humanml3d_llm}, we delve into how different scales and types of LLMs affect the performance of text-based human motion generation tasks on the HumanML3D~\cite{guo2022generating} and KIT-ML~\cite{mandery2015kit} datasets. We observe that: (1) Larger LLMs (\textit{e.g.}, LLaMA 3.1-8B and LLaMA 3-8B) offer distinct advantages over smaller counterparts, achieving significant improvements in fidelity (FID) and multimodal alignment (\textit{i.e.}, R Precision, MultiModal Dist.). To explain, the improved context understanding of LLMs ensures that the output motion aligns closely with intended actions. Further, LLMs with comprehensive world knowledge can synthesize physically plausible motions even when faced with linguistic ambiguity. (2) LLMs fine-tuned by unified instructions demonstrate clear advantage in maintaining semantic consistency and producing motions that are better aligned with textual descriptions. For instance, the fine-tuned Gemma-7B-It outperforms its pre-trained counterpart, achieving a 10\% improvement in R-Precision Top-3 (from 0.673 to 0.715) and a 19\% reduction in FID (from 0.219 to 0.177). (3) Compared to T5-base used in \cite{jiang2023motiongpt,luo2024m}, which requires full fine-tuning, fine-tuning the decoder-only LLaMA 3.1-8B yields a higher R-Precision Top-3 (0.782) and a lower FID (0.191). This indicates that models with more parameters possess capacity to capture complex relationships between text and motion.
% unravel the complex relationship 

\textbf{Consistency with pose control conditions}.
We assess the benefits of pose control by comparing the consistency between the controlled poses and the generated motions on the HumanML3D test sub-set. Concerning each specific task (initial/last/key), motions are generated both with and without pose controls, utilizing the (text+pose)-to-motion and text-to-motion methods, respectively. The results, displayed in Tab.~\ref{tab:pose-consitency} indicate that key-frame consistency is higher with pose controls than in text-only generation counterpart, proving the effectiveness of (text+pose)-to-motion with pose control. Such results highlight the critical role that pose controls play in coherent and contextually appropriate human motion synthesis.

\textbf{Comparison with Separate Training.} We carry out task-specific training on the HumanML3D dataset~\cite{guo2022generating} to test the effectiveness of the proposed MotionGPT-2 model in motion generation. This experimental setup is implemented to determine whether a multi-task learning framework can enhance the performance of each distinct control condition independently. The comparison results are presented in Table~\ref{tab:separate}. Our findings indicate that joint training across all tasks significantly improves performance metrics across the board. Notably, this enhancement is particularly pronounced when utilizing text and the last pose token as input conditions. These results illustrate the value of our multi-modal signals controlled motion generation. MotionGPT-2's ability to generate motions under a specific input condition is strengthened by drawing knowledge from other conditions.

\textbf{Impact of Hierarchical Modeling and Fusion Order.} We conduct comprehensive ablation studies across five backbone models (T5-base, Gemma-2B/7B-It, LLaMA 3-8B, LLaMA 3.1-8B), comparing five fusion strategies: Vanilla VQ-VAE, Body+Hand Concat, Body-First, Alternating, and our proposed Hand-First PA-VQVAE. As shown in Figure~\ref{fig:pa-vqvae-ablation-all-models}, results consistently demonstrate the superiority of the Hand-First design across all backbones and metrics, particularly in semantic alignment (R-Precision, MultiModal Distance), temporal smoothness (FID), and motion diversity. While Body+Hand Concat improves over the vanilla baseline by decoupling representations, hierarchical variants further enhance performance, with Hand-First achieving the best results (e.g., on LLaMA 3.1-8B, Top-1: 0.398, FID: 0.619, MultiModality: 2.821). This ordering effect is theoretically grounded: hand gestures often encode semantic intent that drives subsequent body dynamics. Modeling hands first provides a natural causal prior, yielding more coherent and expressive motion generation. These results collectively validate that the performance gains stem not merely from hierarchical modeling but specifically from the hand-first fusion order.

\vspace{-0.1cm}
\section{Conclusion}
% \subsection{Conclusion}
In this paper, we introduce the MotionGPT-2, a versatile Large Motion-Language Model (LMLM), which can generate and comprehend human motions with general world knowledge of LLMs. Notably, MotionGPT-2 unifies motion-related tasks with multi-modal control signals (\textit{e.g.}, text and single-frame poses) as input by discretizing pose conditions and creating a unified set of instructions from combined textual and pose prompts. By constructing a unified motion-language vocabulary of LLM, we empower the pre-trained LLMs with the ability to integrate the understanding and generation of body kinetics. With well-designed Part-Aware VQVAE, MotionGPT-2 also demonstrates its versatility in addressing the complex 3D whole-body motion generation task, establishing a strong benchmark for researchers. We envision that MotionGPT-2 paves the way for more practical and versatile motion generation systems, offering a fresh perspective in the field.

\vspace{-0.2cm}

{
\bibliographystyle{IEEEtran}
\normalem
\bibliography{motiongpt}
}

% \newpage

\section{Biography Section}
\vspace{-1cm}
\begin{IEEEbiography}[{\includegraphics[width=1in,height=1.25in,clip,keepaspectratio]{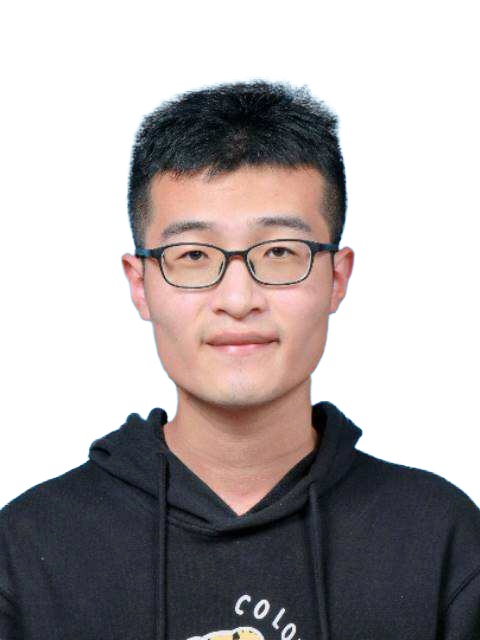}}]{Yuan Wang} received the B.S. degree from Harbin Institute of Technology, Harbin, China and the Master degree from the Institute of Automation, Chinese Academy of Sciences (CASIA). He is currently working toward the Ph.D. degree in the Department of Electronic Engineering, Tsinghua University, Beijing, China. His research interests lie in 3D Computer Vision and Embodied Artificial Intelligence.
\vspace{-1.5cm}
\end{IEEEbiography}

\begin{IEEEbiography}[{\includegraphics[width=1in,height=1.25in,clip,keepaspectratio]{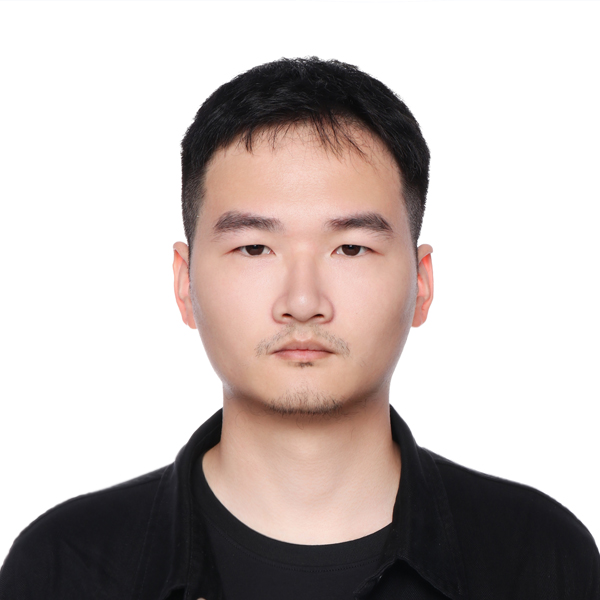}}]{Di Huang} is a Ph.D. student at the University of Sydney. He previously earned his Bachelor’s degree in Biosystems from Zhejiang University. His research interests include 3D vision, particularly in 3D generation and representation learning. He has published approximately 10 papers in top-tier conferences and journals, including CVPR, ICCV, NeurIPS, and ICML.
\vspace{-1.5cm}
\end{IEEEbiography}

\begin{IEEEbiography}[{\includegraphics[width=1in,height=1.75in,clip,keepaspectratio]{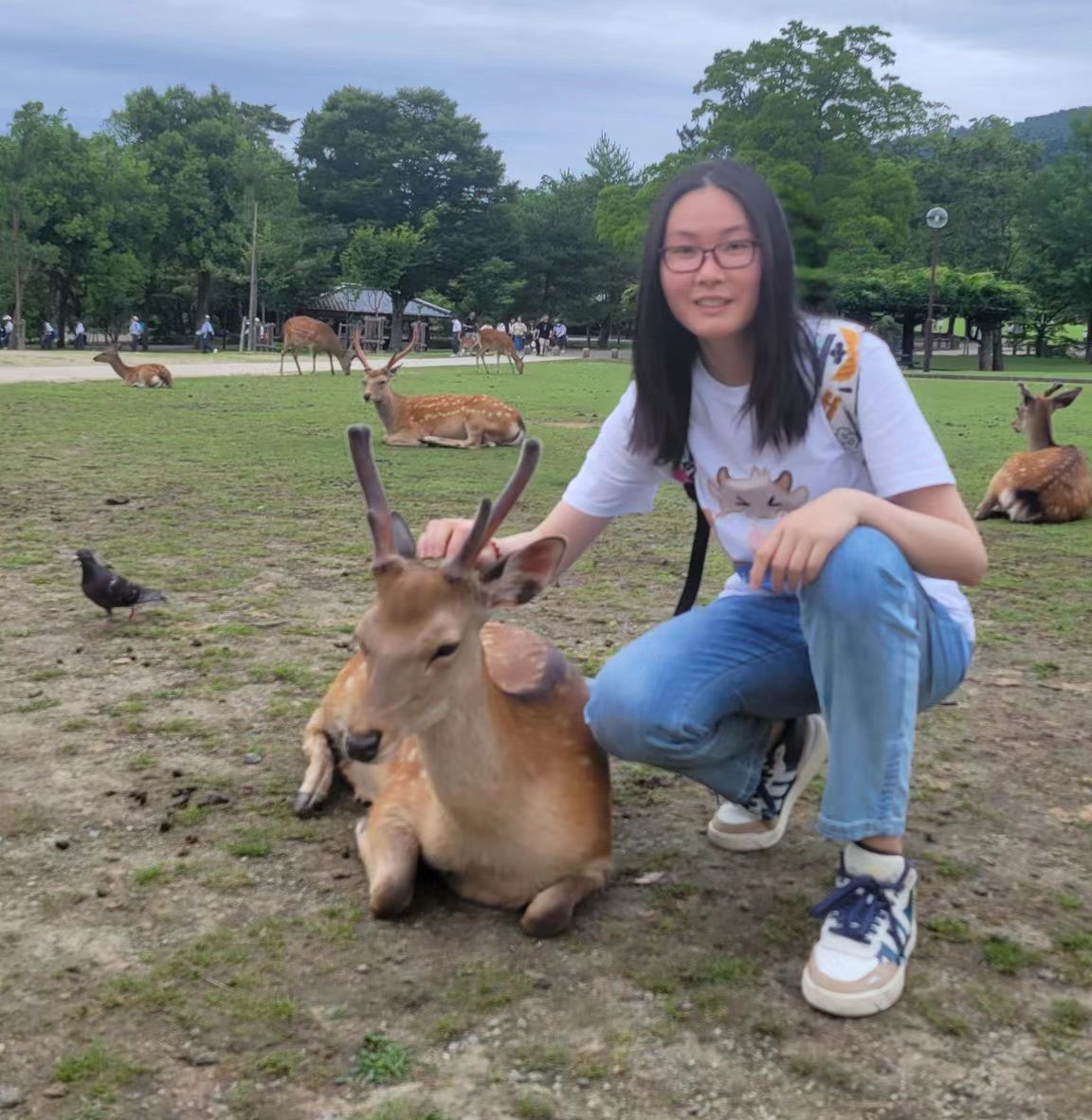}}]{Yaqi Zhang} received her master's degree from the University of Science and Technology of China and bachelor's degree from Xidian University. Her research focuses on 3D human pose and AIGC.
\vspace{-1.4cm}
\end{IEEEbiography}

\begin{IEEEbiography}[{\includegraphics[width=1in,height=1.25in,clip,keepaspectratio]{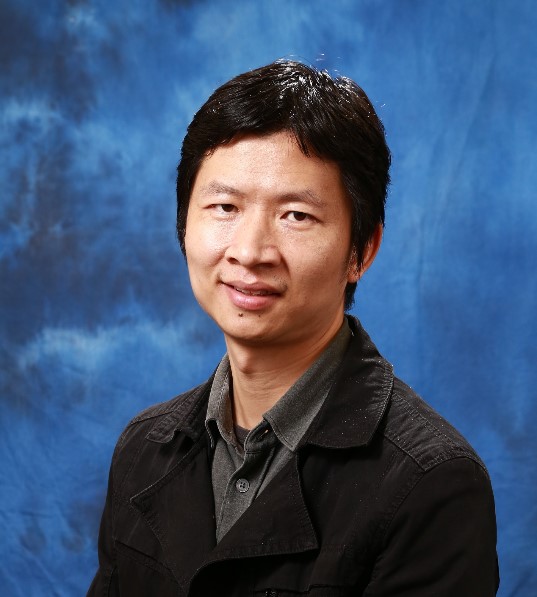}}]{Wanli Ouyang} (Senior Member, IEEE) received
the Ph.D. degree from the Department of Electronic
Engineering, The Chinese University of Hong Kong.
His research interests include deep learning and its
application to computer vision and pattern recognition, image, and video processing. He was awarded
the Australian Research Council Future Fellowship,
meaning that he will be exempted from teaching and
can focus on research in the next four years.
\vspace{-1cm}
\end{IEEEbiography}

\begin{IEEEbiography}[{\includegraphics[width=1in,height=1.25in,clip,keepaspectratio]{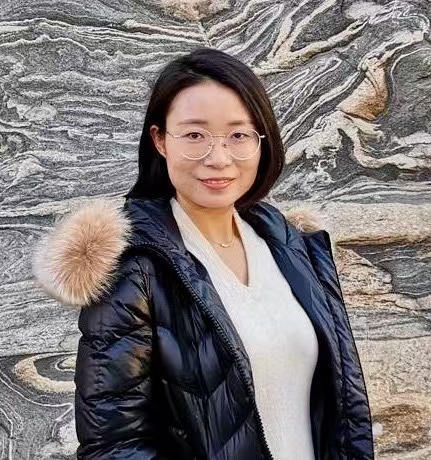}}]{Jile Jiao} received the B.S. degree from Central South University, Changsha, China, in 2009 and the Ph.D. degree in control theory and control engineering from the Institute of Automation, Chinese Academy of Sciences, Beijing, China, in 2014. She is currently an algorithm engineer in Shenxiang Department, Alibaba Group, Beijing, China. She works closely with Tsinghua University. 
Her research interests include computer vision, person reid, multimodal large language models and their application in retail industry.
\vspace{-1cm}
\end{IEEEbiography}

\begin{IEEEbiography}[{\includegraphics[width=1in,height=1.25in,clip,keepaspectratio]{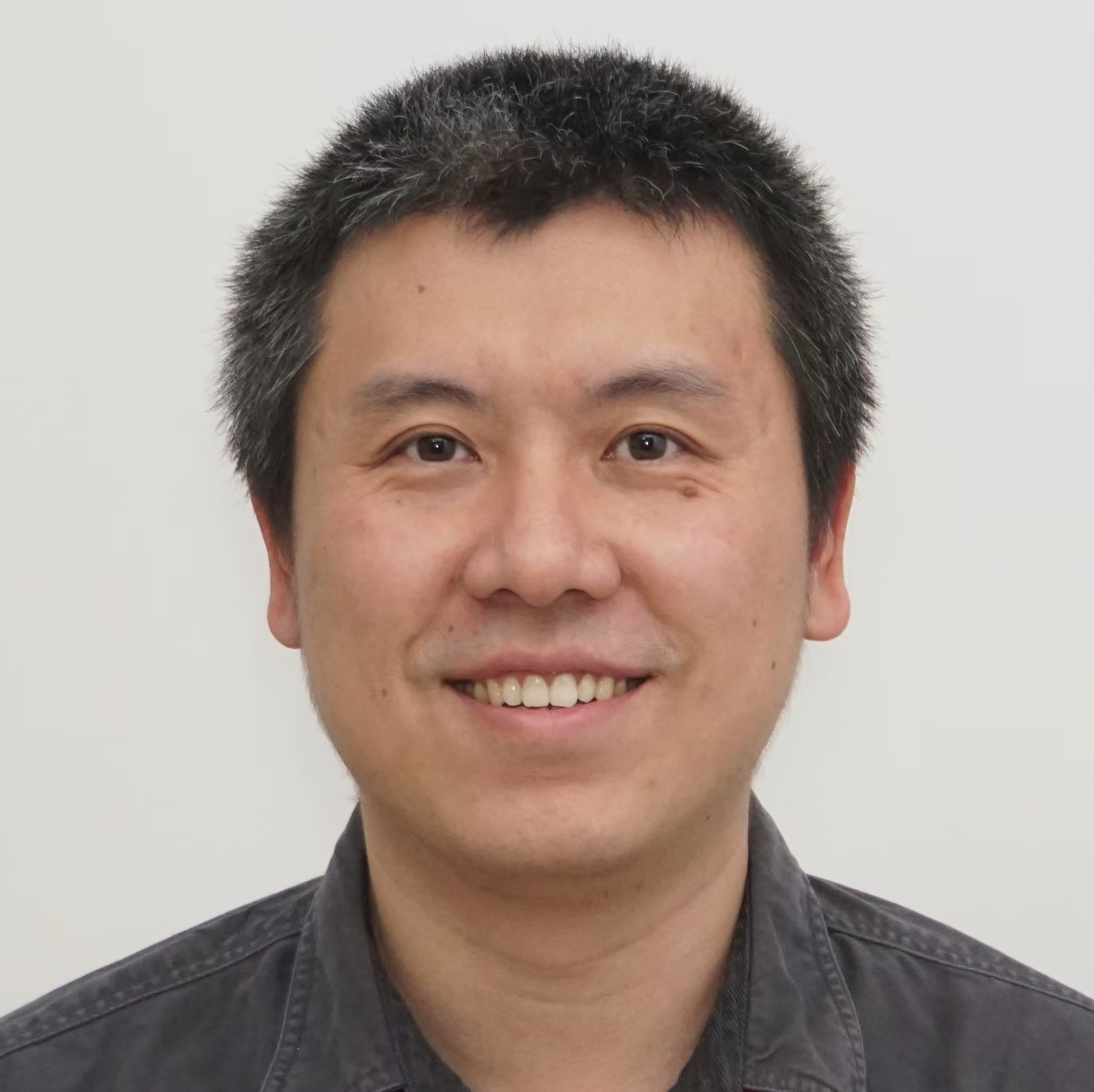}}]{Xuetao Feng} received the B.E. degree from the Department of Automation, Tsinghua University, Beijing, China, in 2004, and the Ph.D. degree in computer application technology from the Institute of Automation, Chinese Academy of Sciences, Beijing, in 2009. He is currently an algorithm engineer in Intime Department, Alibaba Group, Beijing, China. He works closely with Tsinghua University. His research interests include computer vision, image processing, video understanding and their application in e-commerce and retail industry.
\vspace{-1cm}
\end{IEEEbiography}

\begin{IEEEbiography}[{\includegraphics[width=1in,height=1.25in,clip,keepaspectratio]{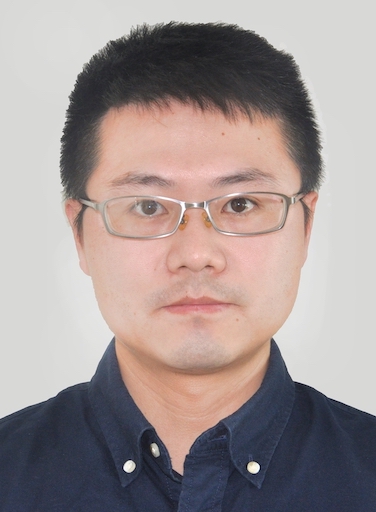}}]{Dan Xu} is an Assistant Professor in the Department of Computer Science and Engineering at HKUST. He was a Postdoctoral Research Fellow in Visual Geometry Group (VGG) at the University of Oxford. He was a Ph.D. in the Department of Computer Science at
the University of Trento. He was also a student research
assistant in MM Lab at the Chinese University of
Hong Kong. He received the best scientific paper
award at ICPR 2016, and a Best Paper Nominee
at ACM MM 2018. He served as Area Chair/Senior PC at multiple main-stream conferences including NeurIPS, CVPR, ECCV, AAAI, ACM Multimedia, WACV, and ACCV.
\vspace{-1cm}
\end{IEEEbiography}

\begin{IEEEbiography}[{\includegraphics[width=1in,height=1.25in,clip,keepaspectratio]{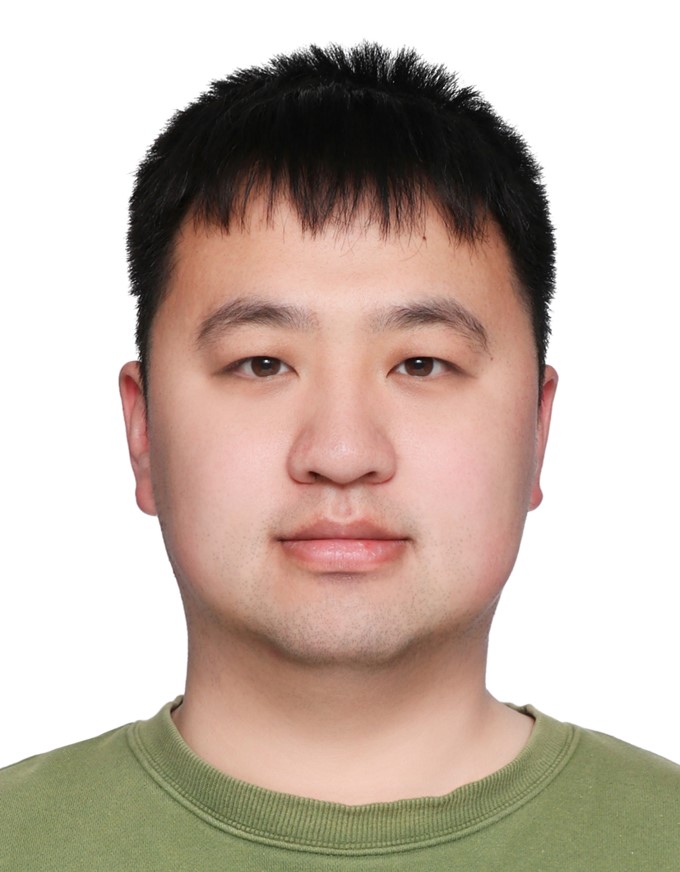}}]{Shixiang Tang} received the Ph.D degree from the University of Sydney. Prior to that, he received the Master of Philosophy from the Chinese University of Hong Kong in 2018 and Bachelor of Science from Fudan University. His interests lie in machine learning and computer vision, especially self-supervised learning and foundation models. He has published about 10 papers in top-tier conferences and journals, e.g., CVPR, ICCV, NeurIPS, Nature Physics and Nature Materials.
\vspace{-1cm}
\end{IEEEbiography}

\vfill

\end{document}